\documentclass[10pt,journal,compsoc]{IEEEtran}
\usepackage{amsmath,amsfonts}
\usepackage{algorithmic}
\usepackage{algorithm}
\usepackage{array}
\usepackage[caption=false,font=normalsize,labelfont=sf,textfont=sf]{subfig}
\usepackage{textcomp}
\usepackage{stfloats}
\usepackage{url}
\usepackage{verbatim}
\usepackage{graphicx}
\usepackage{cite}
\usepackage{graphicx}
\usepackage{xcolor}

\usepackage{amsmath,amsfonts,bm}

\def\eqref#1{equation~\ref{#1}}

\def\1{\bm{1}}

\def\vx{{\bm{x}}}

\DeclareMathAlphabet{\mathsfit}{\encodingdefault}{\sfdefault}{m}{sl}
\SetMathAlphabet{\mathsfit}{bold}{\encodingdefault}{\sfdefault}{bx}{n}

\def\gJ{{\mathcal{J}}}

\def\gS{{\mathcal{S}}}

\def\sR{{\mathbb{R}}}

\usepackage{hyperref}
\usepackage{url}

\usepackage{booktabs}
\usepackage{amsfonts}
\usepackage{nicefrac}
\usepackage{microtype}
\PassOptionsToPackage{prologue,dvipsnames,svgnames,x11names}{xcolor}

\usepackage{wrapfig}
\usepackage{subcaption}

\usepackage{color, colortbl}

\usepackage{algorithm}
\usepackage{listings}
\usepackage{caption}
\usepackage{float}
\definecolor{forestgreen}{rgb}{0, 0.66, 0.31}
\definecolor{DGray}{gray}{0.45}
\definecolor{Gray}{gray}{0.9}
\definecolor{codeblue}{rgb}{0.25,0.5,0.5}
\definecolor{codekw}{rgb}{0.85, 0.18, 0.50}
\definecolor{codekwb}{rgb}{0.0, 0.0, 1.00}
\floatstyle{plain}
\newfloat{Code}{htbp}{Code}
\restylefloat*{Code}
\floatname{Code}{Code}

\usepackage[dvipsnames]{xcolor}

\usepackage[misc]{ifsym}
\usepackage{pifont}
\usepackage{enumitem}
\usepackage{times}
\usepackage{epsfig}
\usepackage{graphicx}
\usepackage{amssymb}

\usepackage{multirow}
\usepackage{booktabs}
\usepackage{float}
\usepackage{bbding}

\usepackage{cleveref}

\newcommand{\eg}{\textit{e.g.}}
\newcommand{\ie}{\textit{i.e.}}

\begin{document}

\crefname{section}{Sec.}{Secs.}
\crefname{table}{Tab.}{Tabs.}
\crefname{figure}{Fig.}{Figs.}
\crefname{equation}{Equ.}{Equs.}

%
\title{SJD++: Improved Speculative Jacobi Decoding for Training-free Acceleration of Discrete Auto-regressive Text-to-Image Generation}
%
%
%
%

\author{
    Yao~Teng, 
    Zhihuan~Jiang, 
    Han~Shi, 
    Xian~Liu, 
    Xuefei~Ning,
    Guohao Dai,~\IEEEmembership{Member,~IEEE,}\\
    Yu~Wang,~\IEEEmembership{Fellow,~IEEE,} 
    Zhenguo~Li, 
    Xihui~Liu\textsuperscript{\Letter},~\IEEEmembership{Member,~IEEE}
    \IEEEcompsocitemizethanks{
        \IEEEcompsocthanksitem Yao~Teng, Zhihuan~Jiang, and Xihui~Liu are with The University of Hong Kong (HKU). (e-mail: tengyao19980325@connect.hku.hk; u3012732@connect.hku.hk; xihuiliu@eee.hku.hk) 
        \IEEEcompsocthanksitem Han~Shi is with Huawei Noah’s Ark Lab. (e-mail: shi.han@huawei.com)
        \IEEEcompsocthanksitem Zhenguo~Li is with Peking University. (e-mail: zhenguol@gmail.com)
        \IEEEcompsocthanksitem Xian~Liu is with xAI. (e-mail: alvinliu0430@gmail.com)
        \IEEEcompsocthanksitem Xuefei~Ning and Yu~Wang are with Tsinghua University. (e-mail: foxdoraame@gmail.com; yu-wang@mail.tsinghua.edu.cn) 
        \IEEEcompsocthanksitem Guohao~Dai is with Shanghai Jiao Tong University. (e-mail: daiguohao@sjtu.edu.cn) 
    }
    \thanks{\Letter~: Corresponding author.} 
}
\markboth{SUBMITTED TO IEEE TRANSACTIONS ON PATTERN ANALYSIS AND MACHINE INTELLIGENCE}%
\ifCLASSOPTIONpeerreview

%



\IEEEtitleabstractindextext{%
\begin{abstract}
Large autoregressive models can generate high-quality, high-resolution images but suffer from slow generation speed, because these models require hundreds to thousands of sequential forward passes for next-token prediction during inference. To accelerate autoregressive text-to-image generation, we propose Speculative Jacobi Decoding++ (SJD++), a training-free probabilistic parallel decoding algorithm. Unlike traditional next-token prediction, SJD++ performs multi-token prediction in each forward pass, drastically reducing generation steps. Specifically, it integrates the iterative multi-token prediction mechanism from Jacobi decoding, with the probabilistic drafting-and-verification mechanism from speculative sampling. More importantly, for further acceleration, SJD++ reuses high-confidence draft tokens after each verification phase instead of resampling them all.
We conduct extensive experiments on several representative autoregressive text-to-image generation models and demonstrate that SJD++ achieves $2\times$ to $3\times$ inference latency reduction and $2\times$ to $7\times$ step compression, while preserving visual quality with no observable degradation.
\end{abstract}

}

\maketitle

\IEEEdisplaynontitleabstractindextext

%
\IEEEpeerreviewmaketitle

\section{Introduction}

\begin{figure}[t]
    \centering
    \includegraphics[width=0.99\linewidth]{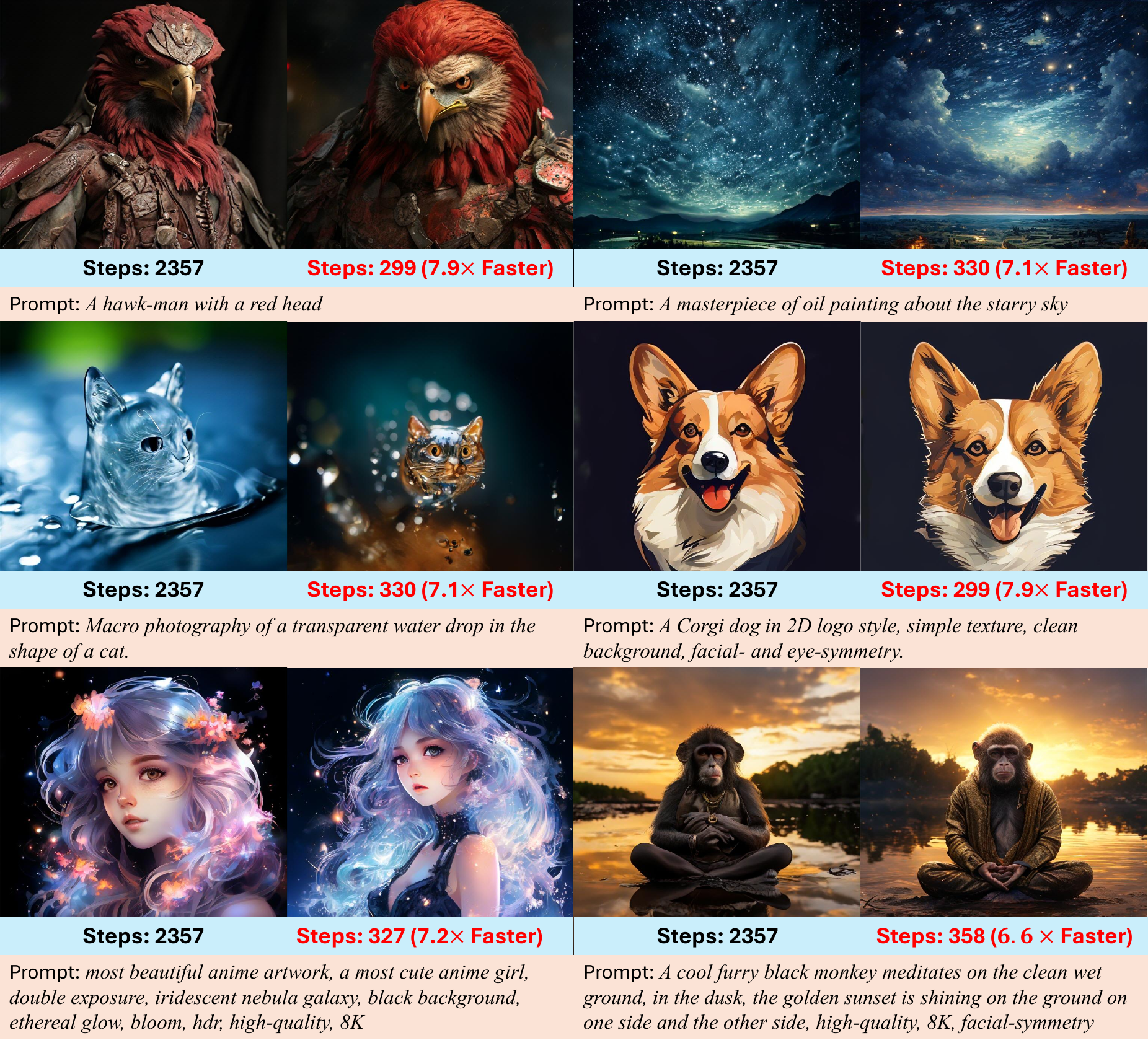}
    \caption{
    We propose Speculative Jacobi Decoding++, a training-free multi-token prediction algorithm, to accelerate autoregressive text-to-image generation by reducing the number of model forward passes (denoted as \texttt{steps}) during inference.
    We perform our algorithm on Lumina-mGPT, and the reduced steps are marked in \textcolor{red}{red}. The original steps are marked in black.
    }
    \label{fig:teaser}
\end{figure}

Auto-regressive (AR) models have emerged as a fundamental framework for generative tasks by predicting the next token conditioned on previously generated ones. 
This paradigm has been extensively adopted across multiple modalities, including language~\cite{gpt4_experiment}, image~\cite{yu2022scaling-parti}, and video generation~\cite{kondratyuk2023videopoet,wang2024loong}. 
In particular, AR text-to-image generation models~\cite{ding2021cogview,dalle,yu2022scaling-parti} have demonstrated remarkable capability in synthesizing high-quality and high-resolution images. 
Due to their strong scalability, these models are regarded as a promising pathway toward developing native multi-modal systems~\cite{team2024chameleon}.
In the paradigm of AR generation, the images are first encoded into sequences of discrete visual tokens through a pre-trained tokenizer with quantizer, and then the transformer-based AR models perform next-token prediction following a conventional raster-scan order.
However, the inherently sequential nature of AR generation imposes substantial inference latency, as generating a single image often requires hundreds or even thousands of token-by-token forward passes. 
While diffusion-based models have been thoroughly investigated for inference acceleration~\cite{consistency_model,luo2023latent-lcm,yin2024one-dmd}, the acceleration of AR text-to-image models has received comparatively limited attention. 
Furthermore, these models typically contain billions of parameters, which makes training-based acceleration techniques such as self-consistency distillation~\cite{kou2024cllms} computationally impractical. 
Consequently, our objective is to accelerate AR text-to-image generation in a completely \emph{training-free} manner.

A natural approach to improve efficiency is to enable AR models to decode multiple tokens in parallel within a single forward pass. 
Early studies on AR image generation explored \emph{Jacobi Decoding}~\cite{ortega2000iterative-original-jacobi}, an iterative parallel decoding strategy that updates token predictions until convergence~\cite{song2021accelerating-jacobi}. 
In each iteration, the model performs one forward pass under a causal attention mask to generate candidate tokens simultaneously, and the process terminates once the sequence stabilizes across consecutive iterations. 
Since the number of iterations required for convergence is generally smaller than the total sequence length, and parallel computation is efficient on GPUs, Jacobi Decoding offers a practical way to reduce inference steps.

Nevertheless, Jacobi Decoding encounters fundamental limitations when applied to contemporary AR text-to-image models. 
Recent architectures~\cite{liu2024lumina-mgpt,chern2024anole,sun2024llamagen} rely heavily on stochastic sampling strategies, such as top-$K$ sampling, to ensure image diversity and fidelity. 
As illustrated in~\cref{fig:res-deterministic-sampling-ar}, increasing the sampling randomness (larger $K$) leads to richer visual details and more diverse outputs, whereas deterministic greedy decoding results in monotonous or even incoherent images. 
The deterministic convergence criterion of Jacobi Decoding is therefore incompatible with such stochastic sampling, rendering it ineffective for accelerating sampling-based AR generation.

To address this challenge, we develop \textbf{Speculative Jacobi Decoding++ (SJD++)} as a probabilistic parallel decoding framework.
SJD++ generalizes the traditional Jacobi iteration into a probabilistic parallel decoding process by integrating the speculative drafting-and-verification mechanism~\cite{leviathan2023fast-speculative,chen2023accelerating-speculative} with a token reuse mechanism for further acceleration. 
In this framework, the model predicts multiple tokens in a single iteration without training, probabilistically accepting a subset of draft tokens according to an acceptance criterion. 
The accepted tokens are appended to the fixed prefix, while the remaining low-confidence draft tokens are resampled and the high-confidence tokens are reused for the next iteration.
In addition, SJD++ incorporates a spatial locality-aware token initialization strategy that exploits the spatial coherence of images to further boost acceleration. 

We evaluate SJD++ on multiple state-of-the-art AR text-to-image generation models, including Lumina-mGPT~\cite{liu2024lumina-mgpt}, Janus-Pro~\cite{chen2025januspro}, SimpleAR~\cite{wang2025simplear}, LlamaGen~\cite{sun2024llamagen}, and Emu3~\cite{Emu-3}, 
across four diverse benchmarks: COCO2017~\cite{coco}, Geneval~\cite{ghosh2023geneval}, PartiPrompt~\cite{yu2022scaling-parti}, and T2I-CompBench++~\cite{t2icompbench++}.
Experimental results consistently demonstrate that SJD++ achieves the state-of-the-art acceleration ratios while preserving image quality across various benchmarks and models. 

Compared with the preliminary conference version~\cite{teng2025accelerating}, this journal paper introduces several non-trivial extensions:
\begin{itemize}
    \item \textbf{Token Reuse Mechanism:} We propose SJD++ that incorporates a new token-reuse mechanism. By retaining high-confidence unaccepted draft tokens across multiple iterations instead of resampling them, SJD++ significantly increases the acceptance rate of draft tokens, leading to substantially greater acceleration.
    This mechanism delivers an additional $1.3\times$ to $3.8\times$ step compression and $1.3\times$ to $1.7\times$ further latency reduction over the original SJD, with no compromise in generation quality.
    \item \textbf{Broader and More Challenging Benchmarks:} We substantially extend the evaluation scope from the original MS-COCO and PartiPrompts to the recent, more comprehensive Geneval and T2I-CompBench++ benchmarks, accompanied by thorough ablation studies on drafting window size and reuse behavior, clearly confirming the benefit of the proposed components.
    \item \textbf{Wider Model Coverage:} Beyond the original base models (Lumina-mGPT, Emu3, and LlamaGen), we further validate SJD++ on a diverse set of modern autoregressive text-to-image models, including SimpleAR and Janus-Pro. Across all tested architectures, SJD++ consistently achieves $2.5\times$ to $7.5\times$ step compression and $2.3\times$ to $3.1\times$ practical latency speedup in a fully training-free setting.
\end{itemize}
These extensions convincingly demonstrate that SJD++ is a universal, high-efficiency, plug-and-play acceleration technique that performs robustly across diverse autoregressive models and evaluation protocols.
\section{Related Work}

\begin{figure}[t]
\centering
\includegraphics[width=0.99\linewidth]{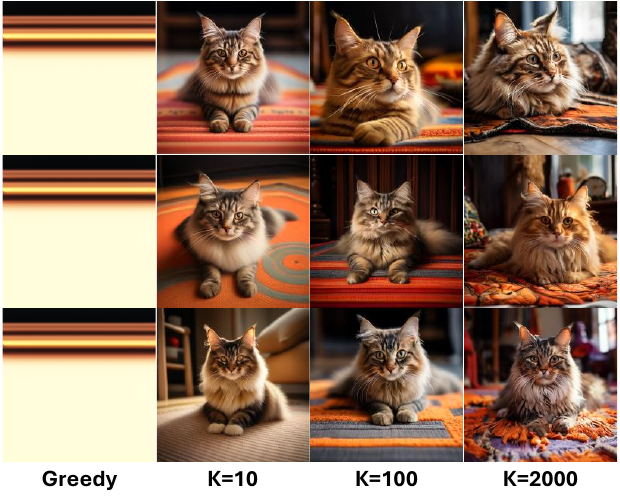}
\caption{Comparison of image diversity under different sampling strategies for Lumina-mGPT~\cite{liu2024lumina-mgpt}. 
Each row shows images generated with the same random seeds using greedy decoding (no randomness), top-$10$, top-$100$, and top-$2000$ sampling.
Greedy decoding produces repetitive and less diverse outputs, while larger $K$ values yield richer textures, colors, and scene variations, highlighting the importance of controlled sampling randomness for high-quality image generation.
}
\label{fig:res-deterministic-sampling-ar}
\end{figure}

\noindent
\textbf{Auto-regressive image generation.}
Auto-regressive (AR) image generation models are characterized by two fundamental properties: \textit{next-token prediction} and \textit{discrete image tokenization}.
Early approaches, such as PixelCNNs~\cite{van2016conditional-pixelcnn,salimans2017pixelcnn++} and PixelSNAIL~\cite{chen2018pixelsnail}, employed convolutional neural networks to model image generation in an autoregressive manner over discretized pixel grids, generating pixels in raster-scan or zigzag order.
DALL-E~\cite{dalle} and CogView~\cite{ding2021cogview} established the modern AR image generation pipeline, where a discrete autoencoder compresses RGB images into image tokens, and a large autoregressive model predicts these tokens sequentially.
Parti~\cite{yu2022scaling-parti} utilizes a Transformer encoder~\cite{transformer} to extract textual embeddings as conditions for image token prediction, achieving text-to-image generation.
LlamaGen~\cite{sun2024llamagen} serves as a class-conditional autoregressive baseline on ImageNet~\cite{imagenet}.
MARS~\cite{he2024mars} extends AR generation to multi-modal settings through a mixture of autoregressive models, where its image branch is initialized from a pre-trained large language model and fine-tuned for image synthesis.
Chameleon~\cite{team2024chameleon} unifies multiple modalities under a shared discrete token space and performs next-token prediction with a single large AR model.
Recent models such as Lumina-mGPT~\cite{liu2024lumina-mgpt} and Anole~\cite{chern2024anole} further fine-tune Chameleon for improved text-to-image generation.

\noindent
\textbf{Acceleration of image generation models.}
The iterative nature of image generation motivates the need for acceleration techniques.
For example, diffusion models—originally trained with over a thousand denoising steps—have been substantially accelerated to perform inference within only a few or dozens of steps.
Given that diffusion models have become the dominant paradigm for text-to-image generation~\cite{Dalle-2,stable_diffusion,sd3}, most prior acceleration methods are designed specifically for them.
Many works reduce the denoising trajectory through various distillation-based strategies~\cite{salimans2022progressive-distill,consistency_model,wang2024phased-cm,kim2023consistency-ctm,xu2024accelerating-sub-path,yin2024one-dmd,yin2024improved-dmd2}, while others focus on optimizing computational complexity~\cite{yuan2024ditfastattn,zhao2024mixdq,ma2024deepcache}.
In contrast, acceleration techniques for AR image generation have received far less attention, mainly due to the historical lack of sufficiently powerful base models. 
Jacobi Decoding was applied to PixelCNNs for inference acceleration in early work~\cite{song2021accelerating-jacobi}, but its deterministic design is incompatible with stochastic token sampling, greatly limiting its effectiveness in modern AR models.
In this paper, we enhance Jacobi Decoding to support stochastic sampling-based decoding, enabling acceleration for large-scale AR text-to-image generators.
It is also noteworthy that each iteration of our decoding process resembles the inference procedure of non-autoregressive models~\cite{chang2022maskgit,tian2024var,li2024autoregressive-mar}.
However, unlike these methods, our approach does not train a separate non-autoregressive model. Instead, it modifies only the inference schedule of existing pre-trained AR models, thus preserving both performance and scalability.

\noindent
\textbf{Acceleration of autoregressive language models.}
The AR paradigm is more extensively explored in natural language processing, where inference acceleration has been pursued from multiple directions.
A large body of work~\cite{zhou2024survey,devoto2024simple-kvcache,liu2024minicache-kvcache,liu2024scissorhands-kvcache,yang2024pyramidinfer-kvcache,deepseekv2-mla,zhang2024pyramidkv,fu2024moa,li2024evaluating} focuses on model compression via pruning, sparsification, quantization, or low-rank factorization, while maintaining the sequential next-token prediction process. 
Some studies aim to parallelize token prediction by fine-tuning models with multiple decoding heads~\cite{gloeckle2024better-multi-token}, though these methods demand extra memory for the additional heads and often require re-training.
Speculative sampling~\cite{leviathan2023fast-speculative,chen2023accelerating-speculative,li2024eagle,sun2024spectr} introduces a lightweight auxiliary model to assist large language models in sequence generation. 
This small model, trained in the same domain, rapidly drafts token sequences, which are then probabilistically verified by the large model in a single forward pass, ensuring consistency with the target conditional distribution.
Jacobi Decoding~\cite{song2021accelerating-jacobi,santilli2023accelerating-jacobi-translation} has also been explored for language generation, enabling iterative multi-token prediction without auxiliary networks.
CLLM~\cite{kou2024cllms} further fine-tunes large language models on Jacobi trajectories to achieve faster decoding, while Lookahead Decoding~\cite{fu2024break-lookahead} adapts training-free Jacobi iterations for language models using an $n$-gram pool obtained through greedy sampling.
In this work, we directly integrate the probabilistic verification mechanism of speculative sampling into Jacobi Decoding, reformulating it into a fully probabilistic algorithm that requires neither auxiliary models nor additional training.

\section{Preliminaries}

\subsection{Auto-regressive Text-to-Image Generation}

Auto-regressive (AR) text-to-image generation models typically consist of three components: 
(1) a discrete image tokenizer that encodes RGB images into compact discrete tokens, 
(2) an autoregressive transformer-based generator that predicts the next image token conditioned on preceding tokens and text prompts, and 
(3) an image decoder that reconstructs pixel-space images from the predicted tokens. 
Among these components, the transformer-based generator is the most computationally expensive, as it performs iterative next-token prediction throughout the inference process. 
Therefore, our work focuses on accelerating the inference of this autoregressive transformer to generate discrete image tokens conditioned on textual inputs.

During each inference step, the transformer predicts a categorical distribution over the tokenizer’s entire vocabulary (implemented via a softmax classifier) and samples one token according to this distribution. 
Formally, given a partially generated token sequence $(x_1, x_2, \ldots, x_i)$, the model parameterized by $\theta$ predicts a probability distribution $p_\theta(x \,|\, \mathbf{x}_{1:i})$, 
where $\mathbf{x}_{1:i}$ denotes the input token sequence, and $x$ represents the random variable corresponding to the next token category. 
A token is then sampled from this distribution as $x_{i+1}$ and appended to the existing sequence $(x_1, x_2, \ldots, x_i)$ for the next decoding step. 
In text-to-image generation, this process begins with a sequence of text tokens followed by a special token that marks the start of image token prediction. 
To encourage diversity and visual richness, top-$K$ sampling is commonly employed as the decoding strategy, where a larger $K$ corresponds to higher sampling randomness.

\subsection{Jacobi Decoding}

Jacobi Decoding formulates the AR inference process as finding the fixed point of a nonlinear triangular system~\cite{song2021accelerating-jacobi}. 
It performs iterative multi-token decoding and can be executed directly on pre-trained AR models without fine-tuning or auxiliary components. 
The decoding procedure for a single token sequence is illustrated in~\cref{fig:jacobi}. 
Given the prefilling or previously decoded tokens, a sequence of candidate tokens is randomly initialized. 
Then, in each Jacobi iteration, the AR model performs one forward pass over all candidate tokens under a causal attention mask. 
The predicted probabilities are converted into tokens via \textit{greedy sampling}, which serve as inputs for the next iteration. 
This iterative process can be mathematically expressed as:
\begin{equation}
x_{i}^{(j+1)} = \arg \max_x p_\theta(x \,|\, \mathbf{x}_{1:i-1}^{(j)}),
\end{equation}
where $i$ denotes the token index and $j$ denotes the iteration index. 
The decoding process continues until convergence is reached, as determined by a deterministic criterion where token values remain unchanged across consecutive iterations.

\noindent
\textbf{Discussion.}
The acceleration of Jacobi Decoding relies on the assumption that multiple tokens can be correctly decoded within a single forward pass during each iteration. 
\cref{fig:jacobi} visualizes this phenomenon: the accepted tokens (\textbf{\textcolor{forestgreen}{green stepped area}}) extend beyond the dashed \textbf{\textcolor{forestgreen}{green triangle outline}}. 
Specifically, after the first Jacobi iteration, the model accepts two consecutive tokens, allowing at least four tokens to be generated with three forward passes. 
In the worst case, only three tokens are produced after three passes~\cite{song2021accelerating-jacobi}, 
which equals the number of steps required by standard autoregressive decoding.

\begin{figure}[t]
\includegraphics[width=1\linewidth]{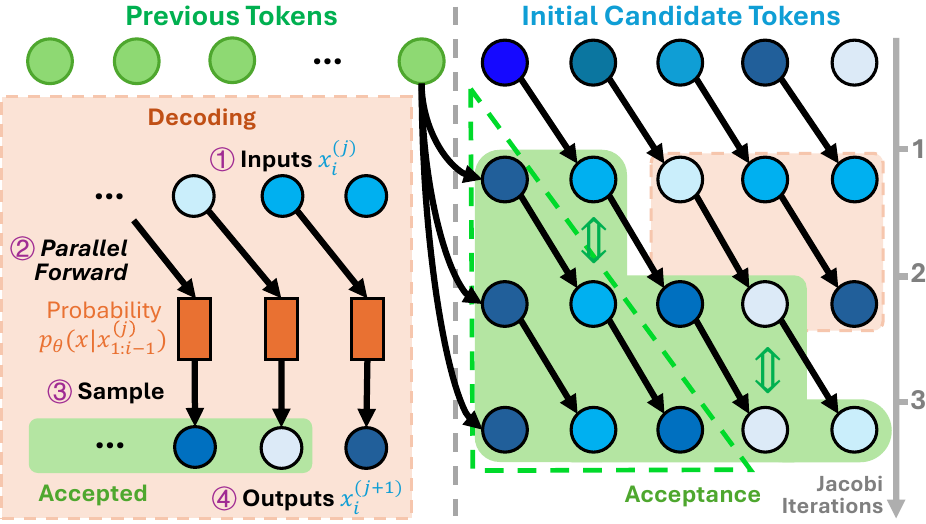}
\caption{The pipeline of the vanilla Jacobi decoding on an autoregressive model. The prediction with sampling is performed in parallel at each Jacobi \textit{iteration}. We use different shades of blue to indicate the differences between the tokens that have not been accepted.}
\label{fig:jacobi}
\end{figure}
\section{Speculative Jacobi Decoding++}

\noindent
\textbf{Analysis.}
The vanilla Jacobi decoding adopts a deterministic convergence criterion, which performs effectively in greedy decoding (no randomness) for language generation~\cite{fu2024break-lookahead,kou2024cllms}. 
However, in autoregressive (AR) text-to-image generation, stochastic decoding plays a crucial role in producing high-quality and diverse images. 
A higher level of randomness leads to more detailed and varied structures, whereas greedy decoding ($K{=}1$) often results in low-quality, repetitive images. 
As illustrated in~\cref{fig:res-deterministic-sampling-ar}, when generating images from the text prompt ``a cat on a mat'', top-$K$ sampling with larger $K$ values yields richer textures and structures, while deterministic decoding produces inferior results. 
Therefore, random sampling-based decoding is essential for text-to-image generation, yet the deterministic convergence rule in Jacobi decoding is incompatible with such stochasticity.

To overcome this limitation, we proposed \textbf{Speculative Jacobi Decoding++ (SJD++)}, a \emph{training-free probabilistic parallel decoding} algorithm inspired by speculative sampling~\cite{leviathan2023fast-speculative,chen2023accelerating-speculative}. 
SJD++ integrates the iterative refinement process of Jacobi decoding with the probabilistic drafting-and-verification mechanism from speculative inference. 
In each iteration, the model decodes multiple tokens in parallel and accepts a subset of them according to a probabilistic acceptance criterion, thereby enabling both parallelism and sampling diversity. 
Furthermore, to reduce the number of decoding iterations, SJD++ introduces a spatially guided initialization strategy that exploits the spatial coherence of images.

\begin{figure*}[t]
    \centering
    \includegraphics[width=0.99\linewidth]{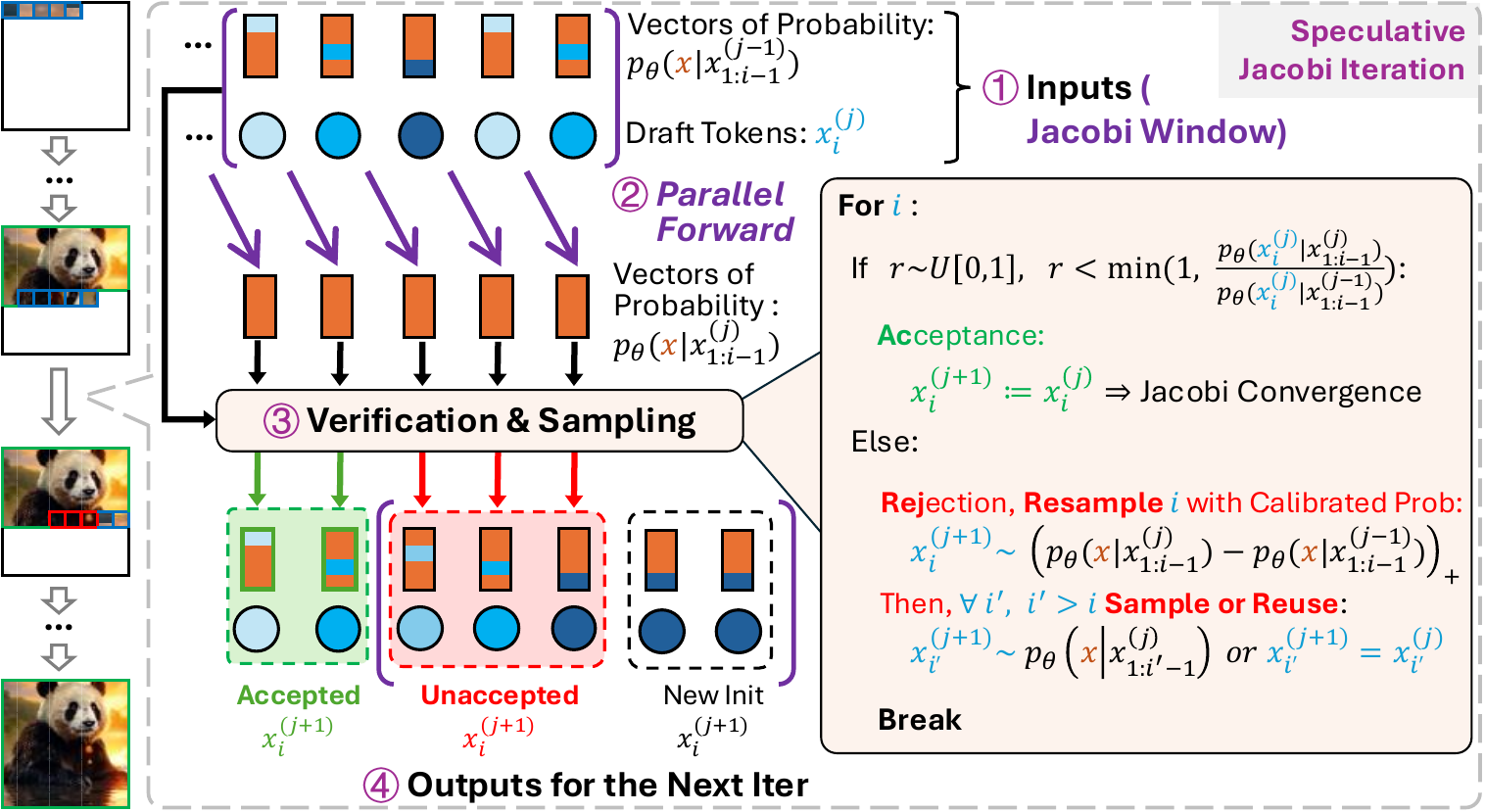}
    \caption{
    \textbf{Overview of one iteration of speculative Jacobi decoding (SJD++).}
    First, a sequence of draft tokens and their corresponding probabilities are provided as input.
    Second, the autoregressive model performs a single forward pass to obtain updated conditional probabilities.
    Third, speculative verification accepts a subset of tokens based on a probabilistic criterion and resamples the remaining ones.
    Finally, accepted tokens are appended to the prefix sequence, while unaccepted tokens, combined with newly initialized tokens, form the next draft sequence for the following iteration.
    }
    \label{fig:speculative-Jacobi-decoding}
\end{figure*}

\subsection{Speculative Jacobi Iteration}
\label{sec:sjd-iter}

After prefilling the text tokens, we generate the image tokens.
Since decoding all tokens simultaneously is computationally expensive, we process multiple tokens in parallel using a sliding \emph{Jacobi window}.
Let $n$ denote the number of accepted tokens, and let $W$ denote the window size.
At iteration $j$, the model takes as input a sequence of draft tokens $(x_{n}^{(j)}, x_{n+1}^{(j)}, \ldots, x_{n+W-1}^{(j)})$, each associated with a conditional probability $p_\theta(x|\mathbf{x}_{1:i-1}^{(j-1)})$ derived from the previous iteration.

\noindent
\textbf{Step 1: Draft token preparation.}
Tokens that remain unaccepted from iteration $(j{-}1)$ are retained, while newly initialized tokens fill the remaining positions to maintain the fixed window size $W$.
The conditional probabilities from the previous iteration serve as the verification reference.

\noindent
\textbf{Step 2: Parallel decoding.}
A single forward pass of the AR model predicts the conditional probabilities $p_\theta(x|\mathbf{x}_{1:i-1}^{(j)})$ for all draft tokens in parallel, with a causal mask in the attention mechanism.

\noindent
\textbf{Step 3: Probabilistic verification.}
We perform speculative verification between two consecutive iterations.
The acceptance of a token $x_i^{(j)}$ is denoted as
\begin{equation}
x_i^{(j+1)} \leftarrow x_i^{(j)} 
\text{ if } r < \min \!\left(1, 
\frac{p_\theta(x_i^{(j)}|\mathbf{x}_{1:i-1}^{(j)})}
{p_\theta(x_i^{(j)}|\mathbf{x}_{1:i-1}^{(j-1)})} \right),
\label{eq:acc-sjdpp}
\end{equation}
where $r \sim \mathcal{U}(0,1)$.
Accepted tokens are appended to the preceding tokens. Thus, for the subsequent iteration $j' > j$, $x_i^{(j')} = x_i^{(j)}$.

\noindent
\textbf{Step 4: Token Resampling and Reusing.}
Once the inequality in~\cref{eq:acc-sjdpp} is not satisfied, 
the remaining tokens are rejected and replaced through a resampling process.
The first unaccepted token 
(\ie, the first position after the accepted prefix) 
is resampled from a \emph{calibrated residual distribution} that captures 
the positive difference between two consecutive predictive probabilities:
\begin{equation}
\tilde{x}_i^{(j+1)} \!\sim\!
\frac{
\max\!\big(0,\; p_\theta(x|\mathbf{x}_{1:i-1}^{(j)}) - p_\theta(x|\mathbf{x}_{1:i-1}^{(j-1)})\big)
}{
\sum_x \max\!\big(0,\; p_\theta(x|\mathbf{x}_{1:i-1}^{(j)}) - p_\theta(x|\mathbf{x}_{1:i-1}^{(j-1)})\big)
}.
\label{eq:rej-sjd}
\end{equation}
This residual sampling focuses the update on tokens whose probabilities 
have significantly increased in the new iteration, 
thereby reducing redundant resampling.
For the subsequent tokens ($i' > i$), SJD++ introduces a 
\textit{confidence-based reuse mechanism} to further avoid unnecessary sampling. 
We compute a confidence mask $C_i^{(j)}$ that measures the local stability 
of token $x_i$ across iterations, defined as the conditional likelihood ratio:
\begin{equation}
C_i^{(j)} =
\frac{
p_\theta\!\left(x_i^{(j)} \mid \mathbf{x}_{1:i-1}^{(j)}\right)
}{
p_\theta\!\left(x_i^{(j)} \mid \mathbf{x}_{1:i-1}^{(j-1)}\right)
}.
\end{equation}
Tokens with high confidence ($C_i^{(j)} > \tau$) are reused directly in the next iteration,
while the others are reinitialized by resampling from the current conditional distribution:
\begin{equation}
x_i^{(j+1)} =
\begin{cases}
x_i^{(j)}, & \text{if } C_i^{(j)} > \tau, \\
\tilde{x}_i^{(j+1)} \!\sim\! p_\theta(x|\mathbf{x}_{1:i-1}^{(j)}), & \text{otherwise}.
\end{cases}
\label{eq:reuse}
\end{equation}
Finally, the accepted and reinitialized tokens are concatenated to form the next Jacobi window.

\subsection{Token Initialization with Spatial Prior}
\label{subsec:initialization}

In the vanilla Jacobi decoding, initial candidate tokens are randomly sampled.
However, natural images exhibit strong \emph{spatial locality}: adjacent regions often share similar textures and semantics.
Leveraging this property can accelerate convergence.
Since AR models typically generate tokens in a raster-scan order (from top-left to bottom-right), we propose several initialization strategies:
(a) repeating the previously generated left-adjacent token,
(b) repeating the previously generated above-adjacent token,
(c) resampling from the predicted distribution of the left-adjacent token, and
(d) resampling from the predicted distribution of the above-adjacent token.
These spatially aware initialization strategies for horizontal-sampling or repeating can achieve additional acceleration in certain structured or repetitive scenes.

\section{Experiments}

In this section, we denote SJD++ without token reuse as SJD. The key difference between SJD and SJD++ lies in the token refinement. For SJD, this process is formulated as:
\begin{equation}
x_i^{(j+1)} \sim p_\theta(x_i \mid \mathbf{x}_{1:i-1}^{(j)}),
\label{eq:refine-sjd}
\end{equation}
which differs from that in \cref{eq:reuse}. Thus, SJD++ can be viewed as an enhanced version of SJD.

\subsection{Implementation Details}
We experiment with four recent and representative autoregressive text-to-image generation models: 
Lumina-mGPT~\cite{liu2024lumina-mgpt}, Janus-Pro~\cite{teng2025accelerating}, SimpleA~\cite{wang2025simplear}, LlamaGen~\cite{sun2024llamagen} and Emu3~\cite{Emu-3}. 
Lumina-mGPT generates $768 \times 768$ images.
Janus-Pro produces $384 \times 384$ images with $24 \times 24$ discrete visual tokens,
while SimpleAR and Emu3 synthesize $1024 \times 1024$ and $720 \times 720$ images, respectively.
LlamaGen generates images with a resolution of $512 \times 512$.
Following the basic configuration of Lumina-mGPT, we set the top-$K$ sampling value to $2000$ to control decoding randomness and the classifier-free guidance weight to $3.0$ for all models. 
Following previous work, we fix $K=2000$ for all models to ensure consistent sampling randomness and fair comparison.  
In all experiments, we enable token reuse only for SJD++ while the standard SJD resamples all tokens at each iteration, and the threshold for reuse is set to $0.5$. 
All evaluations are conducted in a training-free setting without any additional fine-tuning or distillation.

\begin{table*}[t]
\centering
\setlength{\tabcolsep}{5pt}
\caption{
Evaluation on MS-COCO2017~\cite{coco} validation set. 
\texttt{SJD}: Speculative Jacobi Decoding; 
\texttt{SJD++}: SJD with token reuse.
}
\label{tab:coco2017}
\begin{tabular}{l|c|cc|ccc}
\toprule
\multirow{2}{*}{\textbf{Model}} & \multirow{2}{*}{Avg. Latency ($\downarrow$)} & \multicolumn{2}{c|}{Acceleration ($\uparrow$)} & \multirow{2}{*}{FID ($\downarrow$)} & \multirow{2}{*}{HPSv2 ($\uparrow$)} & \multirow{2}{*}{CLIP ($\uparrow$)} \\
 &  & Latency & Step &  &  &  \\
\midrule
\textbf{SimpleAR (0.5B)} & 83.06s & $1.00\times$ & $1.00\times$ & 36.78 & 0.2393 & 28.32 \\
\textit{+ SJD} & 42.81s & $1.94\times$ & $2.33\times$ & 37.00 & 0.2365 & 28.20 \\
\textit{+ SJD++} & \textbf{29.88s} & $\mathbf{2.78\times}$ & $\mathbf{3.28\times}$ & 37.20 & 0.2344 & 28.10 \\
\midrule
\textbf{LlamaGen (0.8B)} & 46.02s & $1.00\times$ & $1.00\times$ & 32.25 & 0.2674 & 30.45 \\
\textit{+ SJD} & 33.18s & $1.39\times$ & $2.10\times$ & 32.40 & 0.2651 & 30.33 \\
\textit{+ SJD++} & \textbf{19.84s} & $\mathbf{2.32\times}$ & $\mathbf{3.26\times}$ & 32.58 & 0.2630 & 30.20 \\
\midrule
\textbf{Janus-Pro (1B)} & 14.50s & $1.00\times$ & $1.00\times$ & 33.97 & 0.2838 & 31.49 \\
\textit{+ SJD} & 8.01s & $1.81\times$ & $1.96\times$ & 34.20 & 0.2819 & 31.38 \\
\textit{+ SJD++} & \textbf{6.29s} & $\mathbf{2.31\times}$ & $\mathbf{2.47\times}$ & 34.40 & 0.2803 & 31.27 \\
\midrule
\textbf{Lumina-mGPT (7B)} & 87.23s & $1.00\times$ & $1.00\times$ & 30.77 & 0.2857 & 31.52 \\
\textit{+ SJD} & 42.73s & $2.04\times$ & $2.22\times$ & 30.90 & 0.2838 & 31.40 \\
\textit{+ SJD++} & \textbf{27.92s} & $\mathbf{3.12\times}$ & $\mathbf{6.44\times}$ & 31.48 & 0.2817 & 31.27 \\
\midrule
\textbf{Emu3-Gen (8B)} & 375.29s & $1.00\times$ & $1.00\times$ & 35.46 & 0.2563 & 29.96 \\
\textit{+ SJD} & 207.61s & $1.81\times$ & $2.01\times$ & 35.60 & 0.2542 & 29.95 \\
\textit{+ SJD++} & \textbf{124.62s} & $\mathbf{3.01\times}$ & $\mathbf{7.50\times}$ & 35.79 & 0.2524 & 29.83 \\
\bottomrule
\end{tabular}
\end{table*}

\begin{table*}[t]
\centering
\setlength{\tabcolsep}{5pt}
\caption{
Evaluation on Geneval~\cite{ghosh2023geneval} benchmark. 
\texttt{SJD}: Speculative Jacobi Decoding; 
\texttt{SJD++}: SJD with token reuse.
}
\label{tab:geneval}
\begin{tabular}{l|c|cc|c|cccccc}
\toprule
\multirow{2}{*}{\textbf{Model}} & 
\multirow{2}{*}{Avg. Latency ($\downarrow$)} & 
\multicolumn{2}{c|}{Acceleration ($\uparrow$)} & 
\multirow{2}{*}{Overall Score ($\uparrow$)} &
\multirow{2}{*}{Counting} & \multirow{2}{*}{Colors} & \multirow{2}{*}{Two} & \multirow{2}{*}{Position} & \multirow{2}{*}{ColorAttri} & \multirow{2}{*}{Single} \\
& & {Latency} & {Step} & &   &   &  &   &  &  \\
\midrule
\textbf{SimpleAR (0.5B)} & 83.55s & $1.00\times$ & $1.00\times$ & 0.5906 
& 0.2984 & 0.8261 & 0.8712 & 0.2547 & 0.3492 & 0.9832 \\
\textit{+ SJD} & 42.73s & $1.96\times$ & $2.34\times$ & 0.5890 
& 0.3011 & 0.8243 & 0.8741 & 0.2493 & 0.3475 & 0.9856 \\
\textit{+ SJD++} & \textbf{29.75s} & $\mathbf{2.81\times}$ & $\mathbf{3.31\times}$ & 0.5994 
& 0.3000 & 0.8298 & 0.8788 & 0.2500 & 0.3500 & 0.9875 \\
\midrule
\textbf{LlamaGen (0.8B)} & 44.73s & $1.00\times$ & $1.00\times$ & 0.5820 
& 0.3125 & 0.8140 & 0.7256 & 0.2163 & 0.2843 & 0.9782 \\
\textit{+ SJD} & 31.92s & $1.40\times$ & $2.12\times$ & 0.5805 
& 0.3137 & 0.8162 & 0.7274 & 0.2140 & 0.2822 & 0.9790 \\
\textit{+ SJD++} & \textbf{19.45s} & $\mathbf{2.30\times}$ & $\mathbf{3.25\times}$ & 0.5887 
& 0.3150 & 0.8185 & 0.7301 & 0.2158 & 0.2850 & 0.9800 \\
\midrule
\textbf{Janus-Pro (1B)} & 14.42s & $1.00\times$ & $1.00\times$ & 0.7300 
& 0.4392 & 0.8867 & 0.8324 & 0.6073 & 0.4981 & 0.9862 \\
\textit{+ SJD} & 7.99s & $1.80\times$ & $1.95\times$ & 0.7100 
& 0.4361 & 0.8819 & 0.8297 & 0.6112 & 0.4995 & 0.9878 \\
\textit{+ SJD++} & \textbf{6.31s} & $\mathbf{2.29\times}$ & $\mathbf{2.46\times}$ & 0.7070 
& 0.4375 & 0.8830 & 0.8283 & 0.6100 & 0.5000 & 0.9875 \\
\midrule
\textbf{Lumina-mGPT (7B)} & 87.19s & $1.00\times$ & $1.00\times$ & 0.4800 
& 0.3581 & 0.7924 & 0.6184 & 0.1037 & 0.1936 & 0.9841 \\
\textit{+ SJD} & 42.76s & $2.04\times$ & $2.23\times$ & 0.4700 
& 0.3594 & 0.7942 & 0.6211 & 0.0981 & 0.1952 & 0.9856 \\
\textit{+ SJD++} & \textbf{27.82s} & $\mathbf{3.13\times}$ & $\mathbf{6.44\times}$ & 0.5124 
& 0.3625 & 0.7979 & 0.6263 & 0.1000 & 0.2000 & 0.9875 \\
\midrule
\textbf{Emu3-Gen (8B)} & 376.01s & $1.00\times$ & $1.00\times$ & 0.4800 
& 0.2342 & 0.8145 & 0.5518 & 0.1137 & 0.1328 & 0.9410 \\
\textit{+ SJD} & 207.27s & $1.81\times$ & $2.01\times$ & 0.4800 
& 0.2361 & 0.8163 & 0.5537 & 0.1115 & 0.1319 & 0.9396 \\
\textit{+ SJD++} & \textbf{124.21s} & $\mathbf{3.03\times}$ & $\mathbf{7.49\times}$ & 0.5092 
& 0.2375 & 0.8191 & 0.5556 & 0.1100 & 0.1300 & 0.9375 \\
\bottomrule
\end{tabular}
\end{table*}

\begin{table*}[t]
\centering
\setlength{\tabcolsep}{5pt}
\caption{
Evaluation on the PartiPrompt~\cite{yu2022scaling-parti} benchmark. 
\texttt{SJD}: Speculative Jacobi Decoding; 
\texttt{SJD++}: SJD with token reuse.
}
\label{tab:parti}
\begin{tabular}{l|c|cc|cc}
\toprule
\multirow{2}{*}{\textbf{Model}} & \multirow{2}{*}{Avg. Latency ($\downarrow$)} & \multicolumn{2}{c|}{Acceleration ($\uparrow$)} & \multirow{2}{*}{HPSv2 ($\uparrow$)} & \multirow{2}{*}{CLIP ($\uparrow$)} \\
 &  & Latency & Step &  &  \\
\midrule
\textbf{SimpleAR (0.5B)} & 83.62s & $1.00\times$ & $1.00\times$ & 0.1912 & 29.17 \\
\textit{+ SJD} & 43.01s & $1.95\times$ & $2.32\times$ & 0.1894 & 29.10 \\
\textit{+ SJD++} & \textbf{29.68s} & $\mathbf{2.82\times}$ & $\mathbf{3.34\times}$ & 0.1871 & 29.00 \\
\midrule
\textbf{LlamaGen (0.8B)} & 45.67s & $1.00\times$ & $1.00\times$ & 0.2681 & 31.42 \\
\textit{+ SJD} & 32.45s & $1.41\times$ & $2.09\times$ & 0.2660 & 31.28 \\
\textit{+ SJD++} & \textbf{19.72s} & $\mathbf{2.32\times}$ & $\mathbf{3.21\times}$ & 0.2638 & 31.15 \\
\midrule
\textbf{Janus-Pro (1B)} & 14.39s & $1.00\times$ & $1.00\times$ & 0.2752 & 32.88 \\
\textit{+ SJD} & 8.05s & $1.79\times$ & $1.87\times$ & 0.2731 & 32.75 \\
\textit{+ SJD++} & \textbf{6.27s} & $\mathbf{2.29\times}$ & $\mathbf{2.48\times}$ & 0.2714 & 32.63 \\
\midrule
\textbf{Lumina-mGPT (7B)} & 87.25s & $1.00\times$ & $1.00\times$ & 0.2913 & 32.28 \\
\textit{+ SJD} & 42.82s & $2.04\times$ & $2.25\times$ & 0.2895 & 32.10 \\
\textit{+ SJD++} & \textbf{27.91s} & $\mathbf{3.12\times}$ & $\mathbf{6.40\times}$ & 0.2873 & 32.07 \\
\midrule
\textbf{Emu3-Gen (8B)} & 375.21s & $1.00\times$ & $1.00\times$ & 0.2164 & 30.42 \\
\textit{+ SJD} & 207.58s & $1.82\times$ & $2.03\times$ & 0.2142 & 30.30 \\
\textit{+ SJD++} & \textbf{124.58s} & $\mathbf{3.01\times}$ & $\mathbf{7.51\times}$ & 0.2123 & 29.83 \\
\bottomrule
\end{tabular}
\end{table*}

\begin{table*}[t]
\centering
\setlength{\tabcolsep}{4pt}
\caption{
Evaluation on T2I-CompBench~\cite{huang2023t2i}. 
\texttt{SJD}: Speculative Jacobi Decoding; 
\texttt{SJD++}: SJD with token reuse.
}
\label{tab:t2icomp}
\begin{tabular}{l|c|cc|ccccccc}
\toprule
\multirow{2}{*}{\textbf{Model}} & \multirow{2}{*}{Avg. Latency ($\downarrow$)} & \multicolumn{2}{c|}{Acceleration ($\uparrow$)} & \multirow{2}{*}{Color} & \multirow{2}{*}{Shape} & \multirow{2}{*}{Texture} & \multirow{2}{*}{2D} & \multirow{2}{*}{3D} & \multirow{2}{*}{CLIP} & \multirow{2}{*}{3-in-1} \\
 &  & Latency & Step &  &  &  &  &  &  &  \\
\midrule
\textbf{SimpleAR (0.5B)} & 83.61s & $1.00\times$ & $1.00\times$ & 0.6402 & 0.3523 & 0.4956 & 0.2058 & 0.3060 & 0.3090 & 0.3558 \\
\textit{+ SJD} & 42.93s & $1.95\times$ & $2.31\times$ & 0.6374 & 0.3505 & 0.4930 & 0.2045 & 0.3038 & 0.3074 & 0.3541 \\
\textit{+ SJD++} & \textbf{29.82s} & $\mathbf{2.81\times}$ & $\mathbf{3.33\times}$ & 0.6346 & 0.3481 & 0.4911 & 0.2033 & 0.3021 & 0.3061 & 0.3527 \\
\midrule
\textbf{LlamaGen (0.8B)} & 44.92s & $1.00\times$ & $1.00\times$ & 0.6014 & 0.3402 & 0.4561 & 0.2037 & 0.2955 & 0.2893 & 0.3398 \\
\textit{+ SJD} & 32.10s & $1.40\times$ & $2.11\times$ & 0.5989 & 0.3384 & 0.4539 & 0.2029 & 0.2936 & 0.2878 & 0.3384 \\
\textit{+ SJD++} & \textbf{19.56s} & $\mathbf{2.30\times}$ & $\mathbf{3.23\times}$ & 0.5965 & 0.3368 & 0.4521 & 0.2020 & 0.2919 & 0.2862 & 0.3370 \\
\midrule
\textbf{Janus-Pro (1B)} & 14.40s & $1.00\times$ & $1.00\times$ & 0.6365 & 0.3532 & 0.4946 & 0.2069 & 0.3076 & 0.3094 & 0.3566 \\
\textit{+ SJD} & 8.03s & $1.80\times$ & $1.86\times$ & 0.6328 & 0.3504 & 0.4924 & 0.2059 & 0.3043 & 0.3072 & 0.3546 \\
\textit{+ SJD++} & \textbf{6.32s} & $\mathbf{2.28\times}$ & $\mathbf{2.47\times}$ & 0.6297 & 0.3484 & 0.4899 & 0.2050 & 0.3010 & 0.3052 & 0.3530 \\
\midrule
\textbf{Lumina-mGPT (7B)} & 87.21s & $1.00\times$ & $1.00\times$ & 0.5628 & 0.3304 & 0.4129 & 0.2006 & 0.2837 & 0.2610 & 0.3179 \\
\textit{+ SJD} & 42.88s & $2.04\times$ & $2.23\times$ & 0.5603 & 0.3293 & 0.4109 & 0.1995 & 0.2821 & 0.2593 & 0.3165 \\
\textit{+ SJD++} & \textbf{27.97s} & $\mathbf{3.13\times}$ & $\mathbf{6.42\times}$ & 0.5578 & 0.3279 & 0.4107 & 0.1996 & 0.2799 & 0.2582 & 0.3163 \\
\midrule
\textbf{Emu3-Gen (8B)} & 374.89s & $1.00\times$ & $1.00\times$ & 0.6549 & 0.3663 & 0.5061 & 0.2066 & 0.3105 & 0.3108 & 0.3607 \\
\textit{+ SJD} & 206.59s & $1.81\times$ & $2.02\times$ & 0.6525 & 0.3628 & 0.5046 & 0.2068 & 0.3089 & 0.3097 & 0.3595 \\
\textit{+ SJD++} & \textbf{124.52s} & $\mathbf{3.01\times}$ & $\mathbf{7.48\times}$ & 0.6508 & 0.3609 & 0.5029 & 0.2071 & 0.3069 & 0.3085 & 0.3583 \\

\bottomrule
\end{tabular}
\end{table*}

\begin{table*}
\centering
\caption{Comparison to other accelerating methods with Lumina-mGPT~\cite{liu2024lumina-mgpt} as baseline  on COCO2017 validation set~\cite{coco}.}
\begin{tabular}{l|ccc}
\toprule
\textbf{Configuration} & Acceleration Latency ($\uparrow$) &  Acceleration Step ($\uparrow$)  & CLIP-Score ($\uparrow$) \\
\midrule
Autoregressive Decoding & 1.00$\times$ & 1.00$\times$ & 31.3 \\
Jacobi Decoding~\cite{song2021accelerating-jacobi} & 1.02$\times$ & 1.04$\times$ & 31.4 \\
EAGLE~\cite{li2024eagle} & 2.10$\times$ & 2.94$\times$ & 33.3 \\
LANTERN~\cite{jang2024lantern} & 2.56$\times$ & 3.63$\times$ & 32.7 \\
ZipAR~\cite{he2024zipar-diagnoal-decode} & 1.82$\times$ & 4.00$\times$ & 31.2 \\
\midrule
{SJD} & 2.05$\times$ & 2.23$\times$ & 31.3 \\
\textbf{SJD++} & \textbf{3.12}$\times$ & \textbf{6.44}$\times$ & 31.3 \\
\bottomrule
\end{tabular}
\label{tab:comp_ac}
\end{table*}

\noindent
\textbf{Metrics.} 
For visual quality evaluation, we report the Fréchet Inception Distance (FID)~\cite{fid}, CLIP-Score~\cite{clip}, and Human Preference Score v2 (HPSv2)~\cite{hpsv2}. 
For quantitative text–image alignment, we additionally report the Geneval~\cite{ghosh2023geneval} score, which measures overall semantic consistency between the prompt and the generated image based on large-scale language–vision models. 
To measure decoding efficiency, we compute the \textit{step compression ratio}~\cite{fu2024break-lookahead}, defined as $\gS = \frac{\text{\# generated tokens}}{\text{\# decoding steps}}$, which reflects the theoretical acceleration ratio. 
For each benchmark, we report both the average step compression ratio and the latency acceleration measured by the wall-clock inference time on a single GPU. 
We further attach the corresponding speedup to the qualitative visualizations for a direct comparison of efficiency and image quality. 
For T2I-CompBench++~\cite{t2icompbench++}, we follow its official evaluation protocol and report sub-metrics covering color, shape, texture, spatial reasoning (2D and 3D), non-spatial alignment (CLIP), and compositional reasoning (3-in-1) scores to comprehensively assess multi-attribute generation performance.

\noindent
\textbf{Benchmark.} 
We evaluate on four widely used text-to-image benchmarks: the validation split of MS-COCO 2017~\cite{coco}, Geneval~\cite{ghosh2023geneval}, PartiPrompt~\cite{yu2022scaling-parti}, and T2I-CompBench++~\cite{t2icompbench++}. 
For COCO2017, we follow the standard evaluation protocol and report FID~\cite{fid}, HPSv2~\cite{hpsv2}, and CLIP-Score~\cite{clip} to jointly assess image fidelity, perceptual realism, and text–image alignment. 
For PartiPrompt~\cite{yu2022scaling-parti}, which provides only textual prompts without ground-truth images, we evaluate using HPSv2 and CLIP-Score to assess perceptual quality and text alignment. 
For Geneval~\cite{ghosh2023geneval}, we adopt its official overall score, which measures semantic consistency between generated images and prompts based on large-scale multimodal language–vision models. 
For T2I-CompBench++~\cite{t2icompbench++}, we follow its official protocol and report sub-metrics including color, shape, texture, spatial reasoning (2D/3D), and compositional reasoning (3-in-1), which comprehensively evaluate multi-attribute and relational understanding. 
Across all benchmarks, we additionally measure latency and the acceleration ratios of both latency and decoding steps to quantify the practical and theoretical speedups achieved by SJD++.

\subsection{Quantitative Results}

As shown in~\cref{tab:coco2017},~\cref{tab:geneval},~\cref{tab:parti}, and~\cref{tab:t2icomp}, 
our speculative Jacobi decoding (SJD) and its enhanced variant SJD++ significantly accelerate the autoregressive text-to-image generation process while maintaining almost identical visual quality across all benchmarks.
Compared with standard autoregressive decoding, SJD already achieves substantial speedups of over $2\times$ on average. 
When further introducing the proposed \textit{token reuse} mechanism, SJD++ brings an additional acceleration without any loss of perceptual or semantic fidelity.

Across all models and datasets, SJD++ consistently improves both latency and step compression.
On average, it reduces inference latency by approximately \textbf{31.5\%} and increases the step compression ratio by around \textbf{1.7×}, compared to SJD. 
Notably, Emu3 achieves the largest overall improvement, with up to \textbf{3.0×} latency acceleration and \textbf{7.5×} step compression on COCO2017. 
Lumina-mGPT and SimpleAR also demonstrate strong efficiency gains, reducing the inference time by about 35–40\%, while Janus-Pro-1B maintains stable acceleration of around 2.3×. 
These results collectively show that the \textit{token reuse} mechanism introduced in SJD++ brings \textbf{remarkably consistent and significant performance improvements} across different model scales and decoding settings.
The Jacobi window length is set to $96$ for Lumina-mGPT~\cite{liu2024lumina-mgpt} and Emu3~\cite{Emu-3}, $32$ for SimpleAR-0.5B~\cite{wang2025simplear} and LlamaGen~\cite{sun2024llamagen}, and $250$ for Janus-Pro-1B~\cite{chen2025januspro}.
Moreover, across all benchmarks (including COCO2017, Geneval, PartiPrompt, and T2I-CompBench) the improvements in efficiency do not compromise generation quality.
The FID, HPSv2, and CLIP scores of SJD++ remain nearly identical to those of the baseline autoregressive decoders, while the Geneval score and compositional metrics on T2I-CompBench are also well preserved. 
This demonstrates that our speculative Jacobi decoding with token reuse achieves substantial acceleration without degrading the model’s visual fidelity, compositional reasoning, or semantic alignment.

\noindent
\textbf{Comparison to existing accelerating methods.} As shown in~\cref{tab:comp_ac}, we compare SJD and SJD++ with the existing speculative/parallel decoding methods, like Eagle~\cite{li2024eagle}, Jacobi Decoding~\cite{song2021accelerating-jacobi}, Lantern~\cite{jang2024lantern}, and ZipAR~\cite{he2024zipar-diagnoal-decode}. We evaluate these methods on the COCO2017 validation set~\cite{coco} and take Lumina-mGPT~\cite{liu2024lumina-mgpt} as the baseline. The results show that our SJD++ achieves superior acceleration while maintaining comparable visual quality.

\noindent
\textbf{Statistics of model outputs.}
We further analyze the statistical properties of token probabilities under both the standard autoregressive decoding and our proposed method. 
Specifically, we compute the mean and standard deviation of the logarithm of token probabilities across all generated image tokens, as shown in~\cref{tab:logit-sta}. 
The results indicate that the tokens accepted by our method exhibit nearly identical statistical behavior to those produced by conventional autoregressive decoding, suggesting that our approach does not erroneously accept tokens with lower confidence.

\begin{figure}[t]
\centering
    \includegraphics[width=0.8\linewidth]{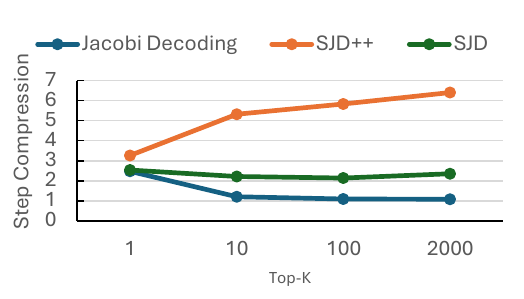}
    \caption{
    SJD++ beats the vanilla Jacobi decoding under various sampling randomness.
    }
    \label{fig:rel-sample-acc}
\end{figure}

\begin{figure}[t]
\centering
    \includegraphics[width=0.8\linewidth]{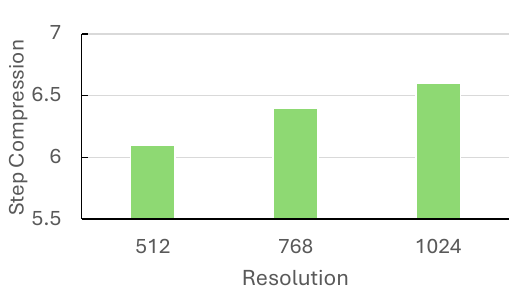}
    \caption{Higher image resolution can result in a slightly larger acceleration in SJD++.}
    \label{fig:reso-speed}
\end{figure}

\begin{figure*}
    \centering
    \includegraphics[width=0.67\linewidth]{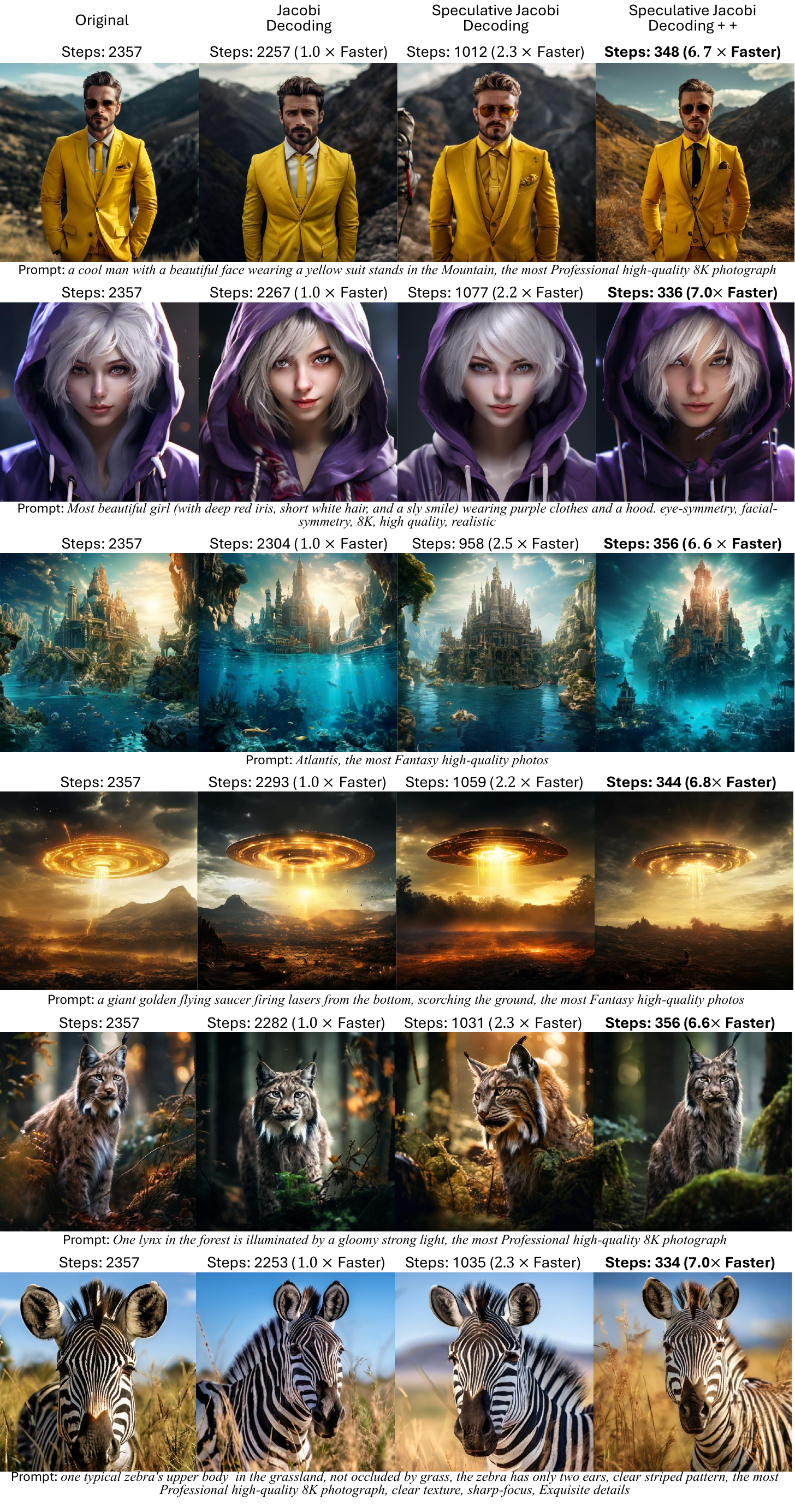}
\caption{
The images generated by Lumina-mGPT~\cite{liu2024lumina-mgpt} with different acceleration methods.
}
\label{fig:lumina-mgpt-sjd}
\end{figure*}

\begin{figure*}
\centering
\includegraphics[width=0.7\linewidth]{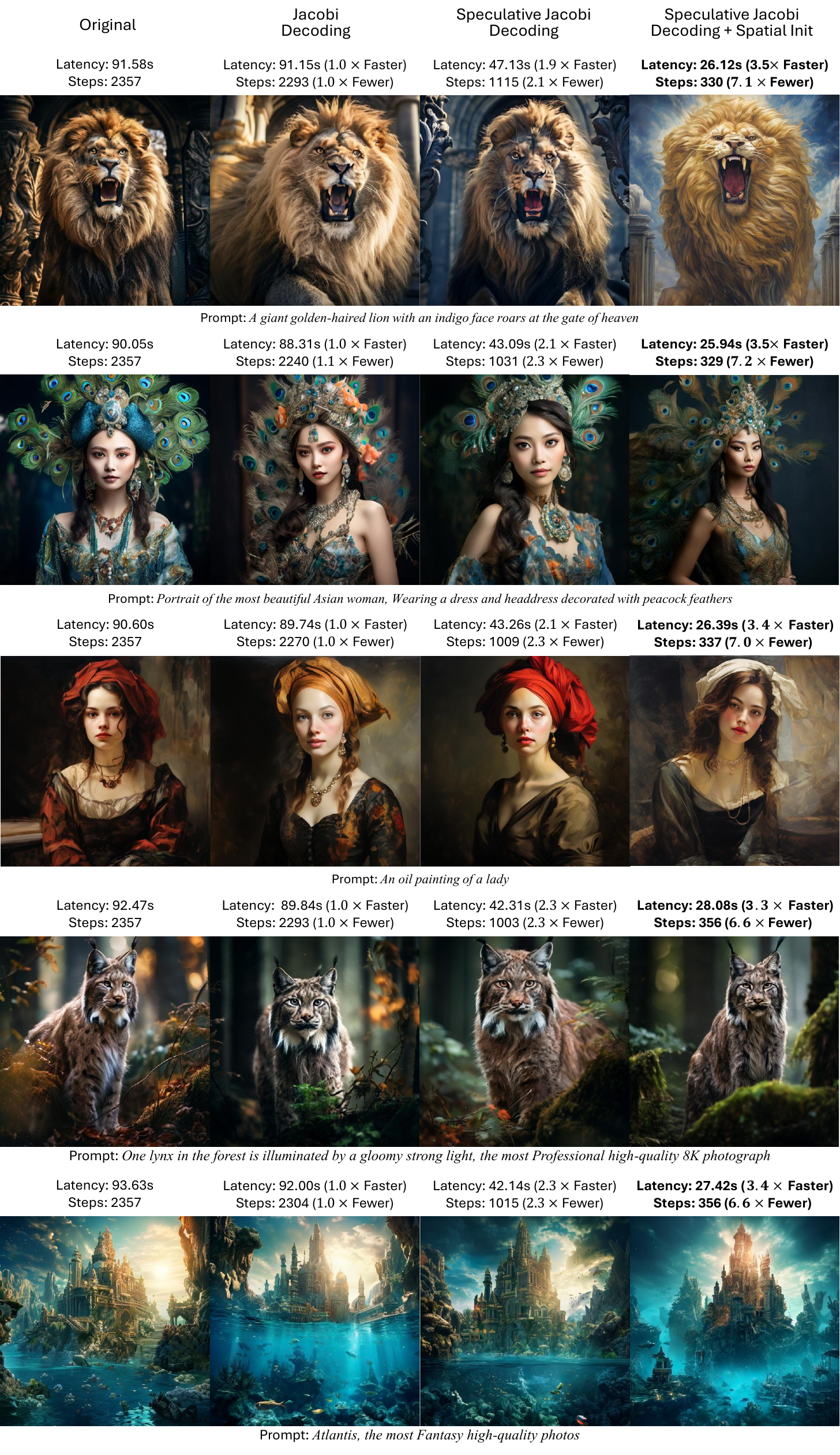}
\caption{
The qualitative comparison of different decoding methods on Lumina-mGPT~\cite{liu2024lumina-mgpt}.}
\label{fig:lumina-mgpt-sjd-comparison-more}
\end{figure*}

\begin{figure*}
    \centering
    \includegraphics[width=0.8\linewidth]{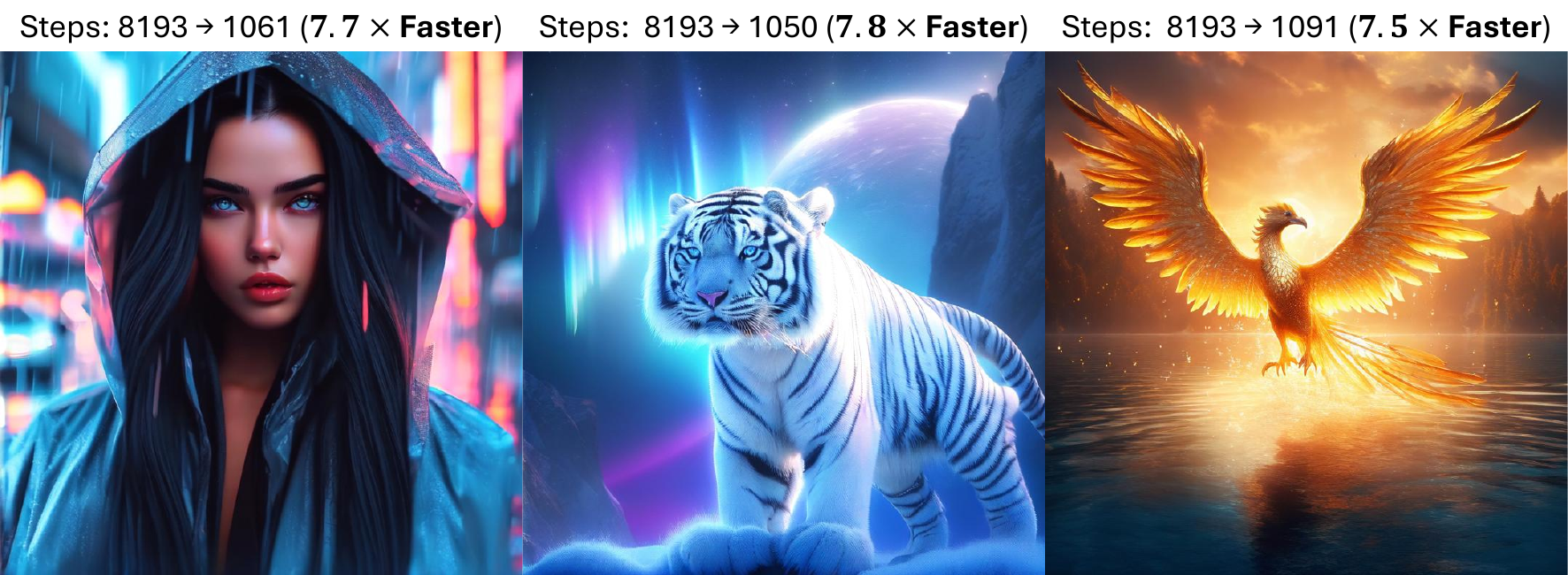}
    \caption{
    The images generated by Emu3~\cite{Emu-3} with SJD++.
    }
    \label{fig:emu-results}
\end{figure*}

\begin{figure*}
    \centering
    \includegraphics[width=0.99\linewidth]{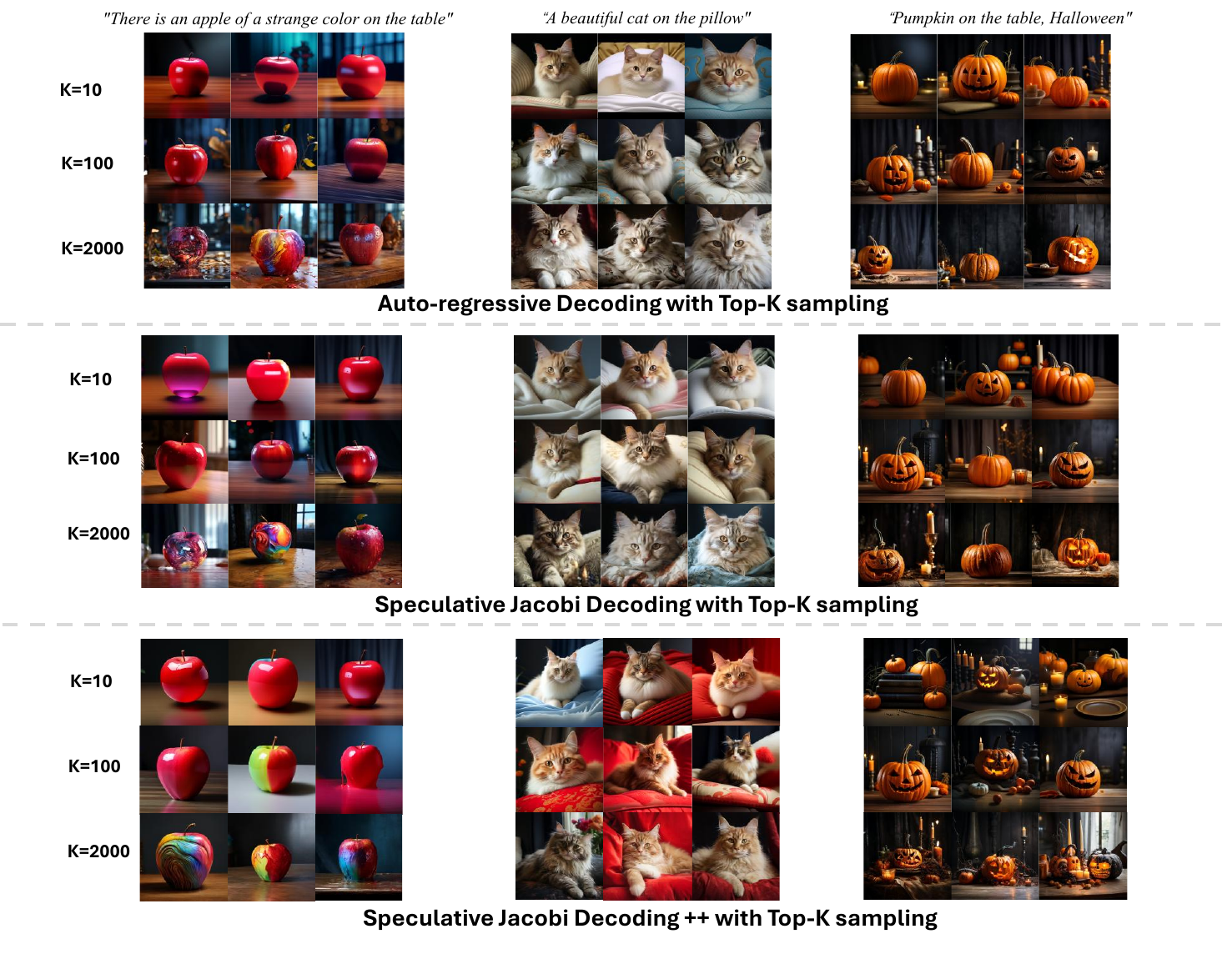}
\caption{
Comparing our methods (SJD and SJD++) to the original autoregressive decoding on the image randomness.
First, considering the random variable in SJD and SJD++, given a column, two images with the same $K$ value cannot be exactly identical.
Second, changing the decoding method from auto-regression to SJD or SJD++ has little influence on the image diversity for each prompt (\eg, given $K=2000$ for each decoding method, the color patterns and styles of the generated apples are similar, and the frequency of the carved faces on pumpkins is also similar).
Third, the top-$K$ sampling still dominates the image randomness regarding texture, color, and local structure details.
Overall, SJD++ achieves comparable image diversity to SJD while offering higher decoding efficiency.
The images in each column share a single random seed.
}
\label{fig:rand-sjd}
\end{figure*}

\begin{figure*}
\centering
\includegraphics[width=0.9\linewidth]{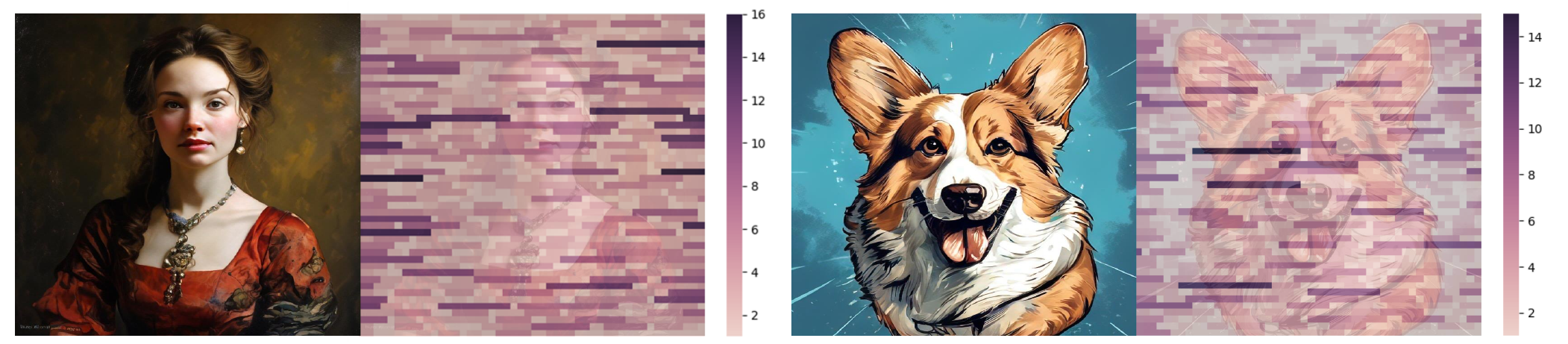}
\caption{
The visualization of the accelerated tokens on 2D space.
}
\label{fig:acc-2d}
\end{figure*}

\subsection{Qualitative Results}

As shown in~\cref{fig:lumina-mgpt-sjd} and~\cref{fig:lumina-mgpt-sjd-comparison-more}, we present images generated by Lumina-mGPT under different decoding configurations, including the original autoregressive decoding, Jacobi decoding, Speculative Jacobi Decoding (SJD), and our improved Speculative Jacobi Decoding++ (SJD++).  
For fair comparison, all generations are initialized with the same random seed.  
We observe that the overall visual quality across all methods remains highly consistent, indicating that both SJD and SJD++ can preserve visual fidelity across diverse image styles and contents.  
Meanwhile, both approaches significantly reduce the number of decoding steps compared to standard autoregressive decoding.  
Specifically, SJD achieves more than $2\times$ step compression, while SJD++ further enhances efficiency through the \textit{token reuse} mechanism, achieving additional acceleration without noticeable degradation in realism or diversity.

Moreover, we extend our method to the recently released large-scale autoregressive model Emu3~\cite{Emu-3}, and observe that SJD++ consistently achieves significant step compression and high-quality generation, as illustrated in~\cref{fig:emu-results}.

\noindent
\textbf{Visualization of acceleration in 2D space.}  
To better understand the spatial distribution of token acceptance, we visualize the acceleration map in~\cref{fig:acc-2d}.  
The color intensity in each region represents the number of accepted tokens, where darker colors indicate longer accepted sequences and hence higher local acceleration.  
We observe that high acceleration typically occurs in texture-smooth background regions, while areas with high-frequency details (such as faces or object boundaries) tend to be updated more conservatively.

\subsection{Ablation Studies}
\label{sec:abla}

We conduct a series of ablation studies to analyze the effectiveness and robustness of our speculative Jacobi decoding (SJD) and its enhanced version (SJD++).
Unless otherwise specified, the experiments are performed on the Lumina-mGPT 7B model, generating images with a resolution of $768 \times 768$. 
We adopt CLIP-Score and the human preference score (HPSv2)~\cite{wu2023human-hpsv2} as the primary metrics for evaluating visual quality.

\noindent
\textbf{Correlation between the sampling strategy and the acceleration ratio.}
We first compare the deterministic Jacobi decoding with our speculative version under various levels of sampling randomness.
As shown in~\cref{fig:rel-sample-acc}, our method maintains stable acceleration across different random sampling schemes and achieves over $2\times$ step compression on average.
In contrast, the deterministic Jacobi decoding can only accelerate greedy decoding (top-$1$ sampling), which is ineffective for text-to-image generation due to its lack of diversity.
Furthermore, as shown in~\cref{tab:topk-clipscore}, the human preference scores remain comparable across different top-$K$ settings, indicating that our method does not compromise visual quality while enhancing decoding efficiency.

\begin{table}[t]
    \centering
    \caption{
    Image quality of various decoding methods on Lumina-mGPT with different top-$K$ values, corresponding to~\cref{fig:rel-sample-acc}.
    Note that the image quality score with greedy sampling is extremely poor, as this setting leads to meaningless images for a lot of prompts (analyzed in~\cref{fig:res-deterministic-sampling-ar}).
    }
    \begin{tabular}{l|l|cc}
    \toprule
    Decoding Methods & Sampling & CLIP-Score & HPSv2\\
    \midrule
    Auto-regression & Top-$1$ & 26.40 & 0.1976 \\
    Auto-regression & Top-$10$ & 32.83 & 0.2950 \\
    Auto-regression & Top-$100$ & 32.41 & 0.3020 \\
    Auto-regression & Top-$2000$ & 32.00 &  0.2965 \\
    \midrule
    Jacobi Decoding & Top-$1$ & 26.34 & 0.1413 \\
    Jacobi Decoding & Top-$10$ & 32.75 & 0.2960 \\
    Jacobi Decoding & Top-$100$ & 32.46 & 
 0.3089 \\
    Jacobi Decoding & Top-$2000$ & 31.68  & 0.3103 \\
    \midrule
    SJD & Top-$1$ & 26.16 & 0.1695 \\
    SJD & Top-$10$ & 32.27 & 0.2942 \\
    SJD & Top-$100$ & 32.65 & 0.2977 \\
    SJD & Top-$2000$ & 31.83 & 0.3020 \\
    \midrule
    SJD++ & Top-$1$ & 26.25 & 0.1583 \\
    SJD++ & Top-$10$ & 32.61 & 0.2958 \\
    SJD++ & Top-$100$ & 32.37 & 0.2982 \\
    SJD++ & Top-$2000$ & 31.72 & 0.3049 \\
    \bottomrule
    \end{tabular}
    \label{tab:topk-clipscore}
\end{table}

\begin{table}[t]
    \centering
    \caption{
    Image quality on Lumina-mGPT with various resolutions. 
    }
    \begin{tabular}{l|c|cc}
    \toprule
    Decoding Methods & Resolutions & CLIP-Score & HPSv2 \\
    \midrule
    Auto-regression & 512 &  29.49 & 0.2503 \\ 
    Auto-regression & 768 & 32.00 & 0.2965 \\ 
    Auto-regression & 1024 & 31.41 & 0.2961 \\ 
    \midrule
    SJD & 512 & 29.69 & 0.2558 \\
    SJD & 768 & 31.83 & 0.3020 \\
    SJD & 1024 & 31.11 & 0.2935 \\ 
    \midrule
    SJD++ & 512 & 29.54 & 0.2563 \\
    SJD++ & 768 & 31.70 & 0.3197 \\
    SJD++ & 1024 & 31.28 & 0.2921 \\ 
    \bottomrule
    \end{tabular}
    \label{tab:resolution-clipscore}
\end{table}

\noindent
\textbf{Influence of image resolution on acceleration.}
We evaluate the 7B Lumina-mGPT model at three resolutions, $512 \times 512$, $768 \times 768$, and $1024 \times 1024$, corresponding to approximately 1K, 2.3K, and 4.1K tokens, respectively.
For each resolution, we compute the average step compression ratio using the same set of prompts.
As shown in~\cref{fig:reso-speed}, our method consistently provides stable acceleration across resolutions, exceeding $2\times$ in all cases. 
Interestingly, the acceleration ratio slightly increases with higher resolution, achieving up to $2.43\times$ at $1024 \times 1024$. 
The visual quality remains consistent across resolutions, as shown in~\cref{tab:resolution-clipscore}.

\noindent
\textbf{Initialization strategies for candidate tokens.}
The initialization of candidate tokens also affects the decoding efficiency.
In scenes with repetitive or spatially correlated patterns, initializing new tokens based on spatial priors tends to converge faster than random initialization.
To examine this, we test on an extreme prompt, \textit{``2D logo of a pure white box on a pure black background''}, and repeat the decoding ten times with different random seeds under each initialization strategy.
The results, presented in~\cref{fig:quali-init}, show that the spatially aware initialization achieves significantly higher step compression than random initialization.
As shown in~\cref{fig:teaser}, simple low-frequency images (\eg, 2D icons) typically require fewer decoding steps than images with complex details.
The visual quality remains similar, as shown in~\cref{tab:init-clipscore}.

\begin{table}[t]
    \centering
    \caption{
    Image quality of SJD on Lumina-mGPT with various token initialization when generating images with simple patterns, corresponding to~\cref{fig:quali-init}.
    }
    \begin{tabular}{l|cc}
    \toprule
    Token Initialization  & CLIP-Score &  HPSv2 \\
    \midrule
    Horizontal Sample &  31.52 & 0.2567 \\
    Vertical Sample &  30.91 & 0.2622 \\
    Horizontal Repeat & 31.17  & 0.2616 \\
    Vertical Repeat & 31.15  & 0.2651 \\
    Random &  31.37 & 0.2681 \\
    \bottomrule
    \end{tabular}
    \label{tab:init-clipscore}
\end{table}

\begin{table}[t]
    \centering
    \caption{
    \centering
    The comparison of token statistics on Lumina-mGPT.
    }
    \begin{tabular}{l|cc}
    \toprule
    \multirow{2}{*}{\begin{tabular}[c]{@{}l@{}} {Decoding Methods} \end{tabular}} & \multicolumn{2}{c}{ Logarithm of Token Probability }  \\
     & Average & Standard Deviation \\
    \midrule
    Auto-regression &  -4.8950 & 2.3457 \\
    SJD &  -4.9007 & 2.3275  \\
    \bottomrule
    \end{tabular}
    \label{tab:logit-sta}
\end{table}

\begin{table}[t]
\centering
\setlength{\tabcolsep}{6pt}
\caption{
Ablation study on the Jacobi window size in SJD++ using Janus-Pro-1B on the Geneval benchmark. 
\texttt{WS}: Jacobi window size. 
NFE: number of function evaluations. 
StepCompr: step compression ratio.
}
\label{tab:ws-ablation}
\begin{tabular}{c|ccc}
\toprule
\textbf{WS} & \textbf{Geneval ($\uparrow$)} & \textbf{NFE ($\downarrow$)} & \textbf{Step Compression ($\downarrow$)} \\
\midrule
16  & 0.7060 & 228.71 & 0.3971 \\
32  & 0.6838 & 198.04 & 0.3438 \\
64  & 0.7225 & 191.51 & 0.3325 \\
\rowcolor{yellow!12}
\textbf{96}  & \textbf{0.7248} & \textbf{189.71} & \textbf{0.3294} \\
128 & 0.7104 & 191.78 & 0.3330 \\
256 & 0.7072 & 192.93 & 0.3349 \\
512 & 0.7093 & 190.79 & 0.3312 \\
\bottomrule
\end{tabular}
\end{table}

\noindent
\textbf{Effect of the Jacobi window size.}
To further investigate the behavior of our enhanced decoding algorithm, we conduct an ablation study on the Jacobi window size in SJD++ using the Janus-Pro-1B model on the Geneval benchmark.
We vary the window size (\texttt{WS}) from 16 to 512 and report the Geneval score, number of function evaluations (NFE), and step compression ratio.
As summarized in~\cref{tab:ws-ablation}, the Geneval score remains relatively stable between 64 and 128, while both excessively small and overly large windows slightly degrade performance.
The best overall trade-off is achieved when \texttt{WS} = 96, yielding the highest Geneval score (0.7248) and the lowest NFE (189.7), corresponding to a step compression ratio of 0.33.
These results suggest that an appropriately sized Jacobi window enables efficient token reuse and stable parallel updates, while too large a window introduces redundant computation and synchronization overhead.

\subsection{Analysis of image randomness}
Like~\cref{fig:res-deterministic-sampling-ar}, we also examine the image randomness with both the autoregressive decoding and our speculative Jacobi decoding. As shown in~\cref{fig:rand-sjd}, first, we find that SJD does introduce some randomness into image generation (the random variable $r$ in~\cref{eq:acc-sjdpp}), so the images generated with autoregressive decoding cannot exactly align with those generated with SJD, even when the random seed is fixed. Therefore, in~\cref{fig:rand-sjd}, given a column, two images with the same $K$ value cannot be exactly identical.
However, in general, the diversity of the set of images is not influenced much.
In~\cref{fig:rand-sjd}, we present the images generated based on three textual prompts. Given the same prompt and $K$ value from top-$K$ sampling, the model with different decoding methods generates images with many similarities. For example, when $K=2000$, for the first prompt ``an apple of a strange color'', the images in the identical columns show the apples with similar color patterns and styles. Also, for the third prompt ``pumpkin on the table'', the frequency of faces carved on the pumpkins is similar for these two decoding methods.
Moreover, the $K$ value in top-$K$ sampling still dominates the image randomness in terms of texture, color, and local structure details. With larger $K$, the image details about textures, colors, and local structures increase. Such image randomness still largely comes from the random token sampling.

\begin{figure*}
\centering
\includegraphics[width=0.8\linewidth]{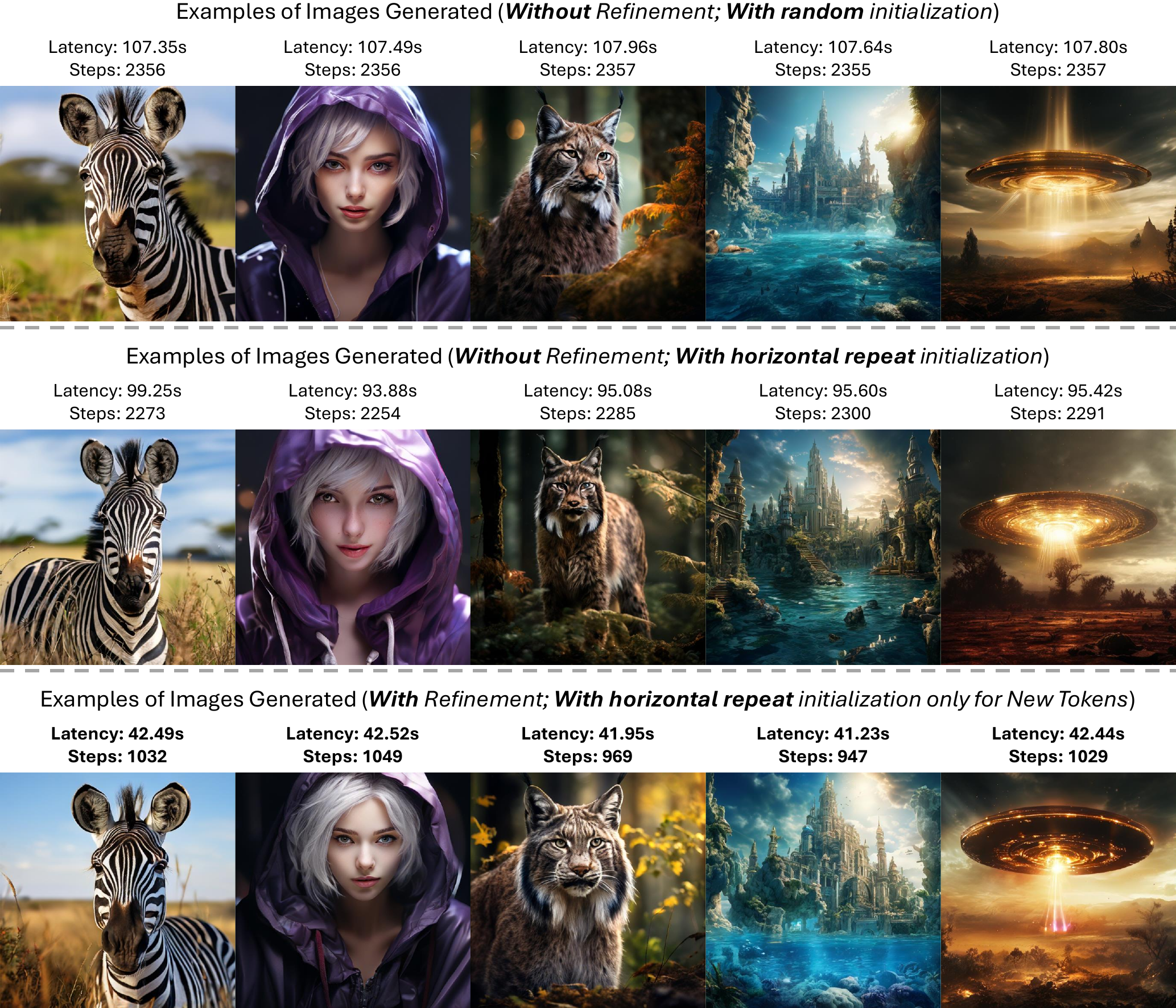}
\caption{
{Ablation studies on acceleration mechanism}:
{examples of images generated by our SJD without or with refining unaccepted tokens.}
When the refinement defined by~\cref{eq:refine-sjd} is \textbf{\textit{NOT}} applied (\ie, using the newly initialized tokens to replace the unaccepted tokens as the draft tokens in the next iterations), there is almost no acceleration (though one of our token initializations with spatial prior, horizontal repeat, can slightly reduce the steps in these images). This illustrates that \textit{\textbf{refining}} unaccepted tokens are essential to the acceleration mechanism in SJD.
}
\label{fig:sjd-no-refinement}
\end{figure*}

\begin{figure}[t]
\centering
\includegraphics[width=1\linewidth]{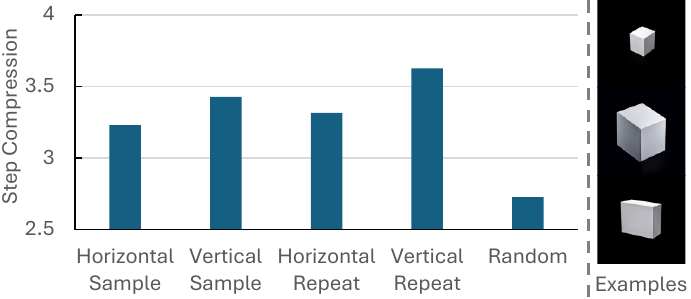}
\caption{
The token initialization strategy impacts the acceleration ratio of image generation that contains simple and repeat patterns (examples of generated images on the right side).
}
\label{fig:quali-init}
\end{figure}

\subsection{Analysis on the refinement}

This section investigates the refinement mechanism of SJD in image generation. 
\textit{Empirically, we find that the acceleration mainly arises from the refinement of unaccepted tokens.} 
Specifically, during the iterative decoding process, some tokens are \textit{continuously resampled}, \ie, their spatial positions in the entire sequence are used across multiple forward passes according to~\cref{eq:refine-sjd}, until they are finally accepted. 
For clarity, we refer to this process of repeatedly resampling a token as \textit{refinement}, following the terminology of fixed-point iteration~\cite{bai2019deep-equilibrium,bai2022deep-deqflow,wang2023deep-deqdet}.

\noindent
\textbf{Acceleration originates from the refinement of unaccepted tokens.} 
During the verification phase of SJD, each token undergoes one of three possible treatments: \textit{acceptance}, \textit{rejection}, or \textit{refinement}, corresponding to~\cref{eq:acc-sjdpp},~\cref{eq:rej-sjd}, and~\cref{eq:refine-sjd}, respectively. 
We empirically observe that relying only on the first two operations (acceptance and rejection) is \textit{\textbf{insufficient}} to achieve acceleration. 
To verify this, we conduct an experiment by disabling the refinement process, \ie, replacing the unaccepted tokens with newly initialized ones in the next iteration instead of reusing them. 
Under this setting, the model requires more than two thousand forward passes to complete an image, compared to roughly one thousand when refinement is enabled. 
Although token initializations with spatial priors (\eg, horizontally repeated tokens) perform slightly better than purely random initialization, their efficiency remains far inferior to that achieved through direct refinement. 
The generated image examples under this condition are shown in~\cref{fig:sjd-no-refinement}. 
These results demonstrate that the acceleration of our method fundamentally arises from the iterative refinement of unaccepted tokens.

\noindent
\textbf{Acceleration through token reuse in SJD++.} 
Although SJD demonstrates theoretical completeness by preserving the token probability to match the predicted distribution of autoregressive models (shown in the appendix), its acceleration ratio remains limited, and its refinement is still under exploration.
Therefore, we propose SJD++ which uses a \textit{token reuse} strategy to enhance decoding efficiency (\ie, replacing \cref{eq:refine-sjd} with \cref{eq:reuse}). 
Instead of resampling all tokens in each verification phase, it selectively reuses high-confidence draft tokens that have already stabilized across iterations, while only resampling uncertain ones. The experimental results in \cref{tab:t2icomp}, \cref{tab:parti}, \cref{tab:coco2017} and \cref{tab:geneval} show that this token reuse strategy greatly improves the acceleration while preserving the visual quality.

\begin{figure*}
    \centering
    \includegraphics[width=0.68\linewidth]{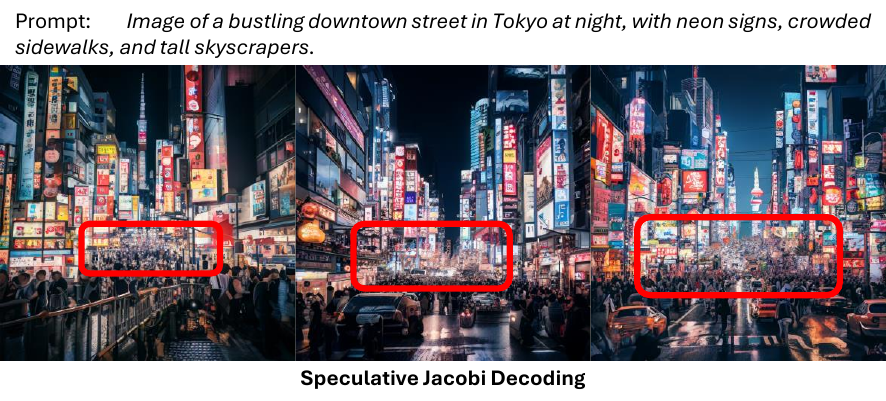}
\caption{
\textbf{Failure Cases}.
In complex image scenarios, our method generates some continuous tokens that result in artifacts, as highlighted by the red boxes. The pre-trained model inaccurately accepts a large sequence of the tokens that cause the artifacts.
}
\label{fig:failure}
\end{figure*}

\subsection{Analysis on Failure Cases}
\label{sec:failure}

As illustrated in~\cref{fig:failure}, when generating images with fine-grained or intricate visual details, both decoding methods may occasionally produce noticeable artifacts. 
Although the autoregressive decoding sometimes yields local distortions, SJD tends to generate continuous sequences of tokens that amplify these artifacts, as indicated by the red boxes in the figure. 
This phenomenon mainly occurs because the pre-trained autoregressive model is not sufficiently robust when handling visually complex scenes, leading it to mistakenly accept a series of draft tokens containing artifacts.

\section{Limitations and Future Work}
\label{sec:limitation}

Our speculative Jacobi decoding and its enhanced variant SJD++ are entirely \textit{training-free} acceleration methods, relying solely on the inference-time resampling and token refinement. 
As a result, the underlying model is not explicitly optimized for multi-token prediction, which limits the theoretical upper bound of acceleration. 
Although SJD++ achieves further improvements through token reuse, it still depends on the stability of the pre-trained transformer architecture and its token-wise consistency.
Future directions include fine-tuning autoregressive models with multi-token consistency objectives, integrating speculative decoding into the training loop, or designing hybrid token schedulers that dynamically adjust reuse ratios.
Beyond image generation, our framework could also extend to long-sequence generation tasks such as video synthesis, where the temporal redundancy between frames provides additional opportunities for parallel refinement and reuse.

\section{Broader Impacts}
\label{sec:broader_impacts}

Image generation models empowered by speculative decoding offer significant benefits to creative industries, education, and human–AI co-creation, enabling faster and more efficient visual synthesis.
However, the same acceleration techniques that improve generation efficiency may also lower the barrier to producing deceptive or harmful visual content.
We encourage both researchers and practitioners to ensure the responsible use of such generative models, including implementing watermarking, content authenticity verification, and transparent model disclosure.
Our work aims to advance the efficiency of generative modeling while maintaining ethical standards and awareness of potential misuse in real-world applications.

\section{Conclusion}
\label{sec:conclusion}

In this paper, we presented \textbf{Speculative Jacobi Decoding++ (SJD++)}, a training-free probabilistic parallel decoding algorithm designed to accelerate auto-regressive text-to-image generation. 
SJD++ introduces a probabilistic convergence mechanism that enables the model to iteratively predict and verify multiple tokens in parallel, significantly reducing the number of decoding steps compared to conventional auto-regressive sampling. 
It incorporates a token reuse strategy that selectively retains high-confidence tokens across iterations for further acceleration. 
Additionally, a spatial-aware token initialization scheme is adopted to exploit the inherent spatial priors of images, facilitating faster convergence in structured visual scenes. 
Comprehensive experiments across multiple representative auto-regressive models and diverse benchmarks demonstrate that SJD++ achieves consistent and substantial acceleration—up to over three times faster decoding—while maintaining comparable visual quality and semantic alignment with the original models. 
We believe that SJD++ provides a general and efficient probabilistic decoding framework that can inspire future research on scalable generation across visual and multimodal domains.


%

\appendix[Proof of the lossless guarantees of SJD]

\label{sec:theorm1}
\noindent
\textbf{Theorem 1} 
The token sampled in each speculative Jacobi iteration satisfies $ p_\theta( x |\vx_{1:i-1}^{(j)})$, where $x$ denotes a token, $j$ denotes the index of iteration, $i$ denotes the token index, and $\theta$ denotes the autoregressive model parameters.

\noindent
\textit{Proof.} 
The main process of speculative Jacobi iteration is decomposed into two cases:
(a) obtaining the token sampled in the previous iteration and then accepting it according to an acceptance probability; 
(b) rejecting the sampled token and resampling a new token according to a calibrated probability.
Thus,
like the proof of the vanilla speculative sampling~\cite{leviathan2023fast-speculative},
to prove the correctness of speculative Jacobi decoding, 
we verify that the conditional probability of a token sampled following the above two cases, alongside the manually designed acceptance and resampling probability, remains $ p_\theta( x |\vx_{1:i-1}^{(j)})$.

For simplicity, by default, we omit the token index $i$ and denote the token category of $\vx_i^{(j)}$ as $x$.
We denote the condition of token $\vx_i^{(j)}$ at the $j$-th Jacobi iteration (\ie, the tokens $\vx^{(j)}_{1:i-1}$ and model weights $\theta$) to $\gJ_j$. Thus, the condition of the $(j-1)$-th Jacobi iteration is denoted as $\gJ_{j-1}$.
Thus, we can denote the probability $ p_\theta( x |\vx_{1:i-1}^{(j)})$ as $p(x | \gJ_j)$, and denote $ p_\theta( x |\vx_{1:i-1}^{(j-1)})$ as $p(x | \gJ_{j-1})$.
We use a random boolean variable $r$ to represent the acceptance.
With these notations, the proof is as follows:

First, the acceptance probability on the token category $x$ is manually set as follows:
\begin{align}
    & p(r \text{ is true} | x , \gJ_j , \gJ_{j-1} ) = \min \{ 1,  \frac{ p(x | \gJ_j) }{ p(x | \gJ_{j-1}) } \},
\label{eq:ac-prob}
\end{align}
and the calibrated resampling probability subsequent to the rejection is set as follows:
\begin{equation}
\begin{aligned}
    & p(x | r \text{ is false} ,  \gJ_j , \gJ_{j-1} ) \\
     =&~ \frac {\max \{ 0, p(x | \gJ_j ) - p(x | \gJ_{j-1} ) \} }{ \sum_{x'} \max \{ 0, p(x' | \gJ_j ) - p(x' | \gJ_{j-1} )  \} }
     .
\label{eq:resample-prob}
\end{aligned}
\end{equation}
Next, we make an assumption that $\gJ_j$ and $x$ are conditionally independent given $\gJ_{j-1}$:
\begin{align}
    & p( \gJ_j | x , \gJ_{j-1}) = p( \gJ_j |  \gJ_{j-1})
\label{eq:independant-j-condition}
\end{align}
This assumption is reasonable due to the properties of the Jacobi iteration and the autoregressive paradigm, \ie, with the observation of the sequence $\vx_{1:i-1}^{(j-1)}$, one of the tokens in $\vx_{1:i-1}^{(j)}$ (denoted as $\vx_{k}^{(j)}$) can be determined by $\vx_{k}^{(j)} = f(\vx_{1:k-1}^{(j-1)}, \theta)~(k < i)$ where the function $f$ indicates the prediction-then-sampling of autoregressive models, so the variable $\vx_i^{(j)}$ is redundant as one of the conditions in the probability $p( \gJ_j | x , \gJ_{j-1})$. Thus,~\cref{eq:independant-j-condition} is reasonable.

Then, with Bayes rule,~\cref{eq:independant-j-condition} has the following equivalence:
\begin{equation}
\begin{aligned}
    & p( \gJ_j | x , \gJ_{j-1}) = p( \gJ_j |  \gJ_{j-1}) \\
   \Leftrightarrow ~~~ & 
   p( x | \gJ_j , \gJ_{j-1}) = p( x |  \gJ_{j-1}) 
\label{eq:j-1-j-condition}
\end{aligned}
\end{equation}

Hence, according to~\cref{eq:ac-prob} and~\cref{eq:j-1-j-condition}, the probability that a token category $x$ is sampled in the previous iteration and subsequently accepted can be computed as:
\begin{equation}
\begin{aligned}
    &p(r \text{ is true} , x | \gJ_j , \gJ_{j-1}) \\
    =&~ p(x | \gJ_{j}, \gJ_{j-1}) \cdot p(r \text{ is true} | x , \gJ_j , \gJ_{j-1}) \\
    =&~ p(x | \gJ_{j-1}) \cdot \min \{ 1,  \frac{ p(x | \gJ_j) }{ p(x | \gJ_{j-1}) } \} \\
    =&~ \min \{   p(x | \gJ_j) ,  p(x | \gJ_{j-1})  \} 
\label{eq:joint-ac}
\end{aligned}
\end{equation}
With~\cref{eq:joint-ac}, we can calculate the probability of rejection with the law of total probability on the token categories: 
\begin{equation}
\begin{aligned}
    &p(r \text{ is false} | \gJ_j , \gJ_{j-1}) \\
    =&~ 1 - p(r \text{ is true} | \gJ_j , \gJ_{j-1}) \\
    =&~ 1 - \sum_{x'} p(r \text{ is true}, x' | \gJ_j , \gJ_{j-1} ) \\
    =&~ \sum_{x'} p(x' | \gJ_j) - \min \{   p(x' | \gJ_j) ,  p(x' | \gJ_{j-1})  \} \\ =&~ \sum_{x'} \max \{ 0, p(x' | \gJ_j) - p(x' | \gJ_{j-1}) \}
    .
\label{eq:demon-rej}
\end{aligned}
\end{equation}
Then, 
with~\cref{eq:resample-prob} and~\cref{eq:demon-rej},
we get the following equation:
\begin{equation}
\begin{aligned}
    &p(x | r \text{ is false} , \gJ_j , \gJ_{j-1} ) \cdot p(r \text{ is false} | \gJ_j , \gJ_{j-1} ) \\
    =& \frac {\max \{ 0, p(x | \gJ_j ) - p(x | \gJ_{j-1} ) \} }{ \sum_{x'} \max \{ 0, p(x' | \gJ_j ) - p(x' | \gJ_{j-1} )  \} } \cdot \sum_{x'} \big\{\\
    &~~~~~~~~~~~\max \{ 0, p(x' | \gJ_j ) - p(x' | \gJ_{j-1} )  \} \big\} \\
    =& \max \{ 0, p(x | \gJ_j ) - p(x | \gJ_{j-1} ) \}
    .
\label{eq:joint-rej}
\end{aligned}
\end{equation}
Since 
\begin{align}
\forall a \in \sR, b \in \sR, ~~ a = \min \{   a ,  b  \} + \max \{ 0, a - b \} ,
\end{align}
we can decompose $p(x | \gJ_j )$ as follows:
\begin{equation}
\begin{aligned}
&p(x | \gJ_j ) \\
=& \min \{   p(x | \gJ_j) ,  p(x | \gJ_{j\!-\!1})  \} \!+\! \max \{ 0, p(x | \gJ_j ) \!-\! p(x | \gJ_{j-1} ) \} .
\label{eq:dempose-cond-prob}
\end{aligned}
\end{equation}
With~\cref{eq:joint-ac},~\cref{eq:joint-rej} and~\cref{eq:dempose-cond-prob}, we can compute:
\begin{equation}
\begin{aligned}
    &p(x | \gJ_j) \\
    =& \min \{   p(x | \gJ_j) ,  p(x | \gJ_{j-1})  \} \!+\! \max \{ 0, p(x | \gJ_j ) \!-\! p(x | \gJ_{j-1} ) \} \\
    =&~ p(x | \gJ_{j-1}) \!\cdot\! p(r \text{ is true} | x , \gJ_j , \gJ_{j-1}) \\
    & ~~~~ \!+\! p(r \text{ is false} | \gJ_j , \gJ_{j-1} ) \!\cdot\! p(x | r \text{ is false} , \gJ_j , \gJ_{j-1} )
    .
    \label{equ:proof-sjd-done}
\end{aligned}
\end{equation}
According to~\cref{equ:proof-sjd-done}, the conditional distribution $p(x | \gJ_j)$ can exactly represent 
(a) obtaining the token sampled in the previous iteration and then accepting it according to an acceptance probability; 
(b) rejecting the sampled token and resampling a new token according to a calibrated probability.
In conclusion, the token sampled in each speculative Jacobi iteration satisfies $ p_\theta( x |\vx_{1:i-1}^{(j)})$.

\ifCLASSOPTIONcaptionsoff
  \newpage
\fi



\bibliographystyle{IEEEtran}
\bibliography{iclr2025_conference}
%



%




\end{document}